\documentclass[11pt,a4paper]{article}
\pdfoutput=1
\usepackage{iclr2020_conference,times}
\usepackage{times}
\usepackage{latexsym}
\usepackage{multirow}
\usepackage{hyperref}
\usepackage{booktabs}
\usepackage{graphicx}
\usepackage{amsmath}
\usepackage{tablefootnote}
\usepackage{xspace}
\usepackage{dsfont}
\usepackage{balance}
\usepackage{url}
\usepackage[T1]{fontenc}
\usepackage[utf8]{inputenc}
\usepackage{fixltx2e}

\newcommand{\gianttablefont}{\fontsize{7}{7}\selectfont}

\usepackage{titlesec}
\titlespacing{\paragraph} {0pt}{0.2ex plus 0.1ex minus .2ex}{1em}

\newcommand{\nlstring}[1]{\emph{#1}}

\newcommand{\sentref}{x}
\newcommand{\senthyp}{\hat{x}}

\newcommand{\reals}{\mathds{R}}

\newcommand{\bert}{\textsc{BERT}\xspace}
\newcommand{\bertbase}{\textsc{BERT}_{{\rm base}}\xspace}
\newcommand{\bertchinese}{\textsc{BERT}_{{\rm chinese}}\xspace}
\newcommand{\bertmulti}{\textsc{BERT}_{{\rm multi}}\xspace}
\newcommand{\roberta}{\text{RoBERTa}_{{\rm large}}\xspace}

\newcommand{\metric}[1]{\textsc{#1}\xspace}
\newcommand{\bleu}{\metric{Bleu}}
\newcommand{\sentbleu}{\metric{SentBleu}}
\newcommand{\meteor}{\metric{Meteor}}
\newcommand{\rouge}{\metric{Rouge}}
\newcommand{\spice}{\metric{Spice}}
\newcommand{\blend}{\metric{Blend}}
\newcommand{\ruse}{\metric{Ruse}}
\newcommand{\beer}{\metric{Beer}}
\newcommand{\meant}{\metric{Meant}}

\newcommand{\leic}{\metric{Leic}}

\newcommand{\method}{\metric{BERTScore}}
\newcommand{\moverscore}{\metric{MoverScore}}

\newcommand{\idf}{{\rm idf}}
\newcommand{\tf}{{\rm tf}}

\newcommand{\methodp}{P_\bert}
\newcommand{\methodr}{R_\bert}
\newcommand{\methodf}{F_\bert}
\newcommand{\hatmethodp}{\hat{P}_\bert}
\newcommand{\hatmethodr}{\hat{R}_\bert}
\newcommand{\hatmethodf}{\hat{F}_\bert}

\newcommand{\exactp}{\text{Exact-P}}
\newcommand{\exactpn}{\exactp_n}
\newcommand{\exactr}{\text{Exact-R}}
\newcommand{\exactrn}{\exactr_n}
\usepackage{amsmath,amsfonts,bm}

\def\eqref#1{equation~\ref{#1}}
\def\1{\bm{1}}

\DeclareMathAlphabet{\mathsfit}{\encodingdefault}{\sfdefault}{m}{sl}
\SetMathAlphabet{\mathsfit}{bold}{\encodingdefault}{\sfdefault}{bx}{n}

\title{\method: Evaluating Text Generation with BERT}

\author{Tianyi Zhang\footnotemark[1]\hspace{4pt}\footnotemark[2]\hspace{4pt}$^\ddagger$$^\diamond$\hspace{-3pt},\hspace{1pt} Varsha Kishore\footnotemark[1]\hspace{4pt}$^\ddagger$\hspace{-3pt}, \hspace{1pt}Felix Wu\footnotemark[1]\hspace{4pt}$^\ddagger$\hspace{-3pt}, \hspace{1pt}Kilian Q. Weinberger\footnotemark[2]\hspace{4pt}$^\ddagger$$^\diamond$\hspace{-3pt}, {\normalfont and} Yoav Artzi$^\ddagger$$^\mathsection$
\vspace{0.5em}\\
$^\ddagger$Department of Computer Science and $^\mathsection$Cornell Tech, Cornell University\\
 {\tt \{vk352, fw245, kilian\}@cornell.edu \hspace{1em} \{yoav\}@cs.cornell.edu}
 \vspace{0.5em}\\
 $^\diamond$ASAPP Inc.\\
 {\tt tzhang@asapp.com}}

\date{}

\iclrfinalcopy % Uncomment for camera-ready version, but NOT for submission.
\begin{document}
\maketitle
\begin{abstract}
  We propose \textit{\method{}}, an automatic evaluation metric for text generation. Analogously to common metrics, \method computes a similarity score for each token in the candidate sentence with each token in the reference sentence. However, instead of exact matches, we compute token similarity using contextual embeddings. We evaluate using the outputs of 363 machine translation and image captioning systems. \method correlates better with human judgments and provides stronger model selection performance than existing metrics. Finally, we use an adversarial paraphrase detection task to show that \method is more robust to challenging examples when compared to existing metrics. 
\renewcommand{\thefootnote}{\fnsymbol{footnote}}
\footnotetext[1]{Equal contribution. $^\dagger$ Work done at Cornell.}
\renewcommand*{\thefootnote}{\arabic{footnote}}

\end{abstract}

\section{Introduction}
\label{sec:intro}

Automatic evaluation of natural language generation, for example in machine translation and caption generation, requires comparing candidate sentences to annotated references. 
The goal is to evaluate semantic equivalence. 
However, commonly used methods rely on surface-form similarity only. 
For example, \bleu~\citep{bleu}, the most common machine translation metric, simply counts $n$-gram overlap between the candidate and the reference. 
While this provides a simple and general measure, it fails to account for meaning-preserving lexical and compositional diversity.

In this paper, we introduce \method, a language generation evaluation metric based on pre-trained \bert contextual embeddings~\citep{bert}.
\method computes the similarity of two sentences as a sum of cosine similarities between their tokens' embeddings.

\method addresses two common pitfalls in $n$-gram-based metrics~\citep{meteor}. 
First, such methods often fail to robustly match paraphrases. 
For example, given the reference \nlstring{people like foreign cars}, \bleu and  \meteor~\citep{meteor} incorrectly give a higher score to \nlstring{people like visiting places abroad}  compared to \nlstring{consumers prefer imported cars}. 
This leads to performance underestimation  when semantically-correct phrases are penalized because they differ from the surface form of the reference. 
In contrast to string matching (e.g., in \bleu) or matching heuristics (e.g., in \meteor), we compute similarity using contextualized token embeddings, which have been shown to be effective for paraphrase detection~\citep{bert}. 
Second, $n$-gram models fail to capture distant dependencies and penalize semantically-critical ordering changes~\citep{Isozaki10:autoeval}. 
For example, given a small window of size two, \bleu will only mildly penalize swapping of cause and effect clauses (e.g. \nlstring{A because B} instead of \nlstring{B because A}), especially when the arguments A and B are long phrases. 
In contrast, contextualized embeddings are trained to effectively capture distant dependencies and ordering.

We experiment with \method on machine translation and image captioning tasks using the outputs of 363 systems by correlating \method and related metrics to available human judgments. 
Our experiments demonstrate that \method correlates highly with human evaluations. 
In machine translation, \method shows stronger system-level and segment-level correlations with human judgments than existing metrics on multiple common benchmarks and demonstrates strong model selection performance compared to \bleu. 
We also show that \method is well-correlated with human annotators for image captioning, surpassing \spice, a popular task-specific metric~\citep{spice}. 
Finally, we test the robustness of \method{} on the adversarial paraphrase dataset PAWS~\citep{paws}, and show that it is more robust to adversarial examples than other metrics. 
The code for \method is available at  \href{https://github.com/Tiiiger/bert_score}{\tt https://github.com/Tiiiger/bert\_score}.

\section{Problem Statement and Prior Metrics}
\label{sec:related_works}

Natural language text generation is commonly evaluated using annotated reference sentences. 
Given a reference sentence $\sentref{}$  tokenized to $k$ tokens $\langle \sentref{}_1, \dots, \sentref{}_k  \rangle$ and a candidate  $\senthyp{}$ tokenized to $l$ tokens $\langle \senthyp_1, \dots, \senthyp_l\rangle$, a generation evaluation metric is a function $f(\sentref{}, \senthyp{}) \in \reals$. 
Better metrics have a higher correlation with human judgments. 
Existing metrics can be broadly categorized into using $n$-gram matching, edit distance, embedding matching, or learned functions. 

\subsection{$n$-gram Matching Approaches}

The most commonly used metrics for generation count the number of $n$-grams that occur in the reference $\sentref{}$ and candidate $\senthyp{}$. The higher the $n$ is, the more the metric is able to capture word order, but it also becomes more restrictive and constrained to the exact form of the reference. 

Formally, let $S^n_{\sentref{}}$ and $S^n_{\senthyp{}}$ be the lists of token $n$-grams ($n \in \mathbb{Z}_+$) in the reference $\sentref{}$ and candidate $\senthyp{}$ sentences. 
The number of matched $n$-grams is $\sum_{w \in S^n_{\senthyp{}}} \mathbb{I}[w \in S^n_{\sentref{}}]$, 
where $\mathbb{I}[\cdot]$ is an indicator function. 
The exact match precision ($\exactpn$) and recall ($\exactrn$) scores are:
\begin{eqnarray*}
\exactpn = \frac{\sum_{w \in S^n_{\senthyp{}}} \mathbb{I}[w \in S^n_{\sentref{}}]}{\left \lvert S^{n}_{\senthyp{}} \right \rvert} \quad \text{and} \quad
\exactrn = \frac{\sum_{w \in S^n_{\sentref{}}} \mathbb{I}[w \in S^n_{\senthyp{}}]}{\left \lvert S^{n}_{\sentref{}} \right \rvert}.
\end{eqnarray*}
Several popular metrics build upon one or both of these exact matching scores.

\paragraph{\bleu}
The most widely used metric in machine translation is \bleu~\citep{bleu}, which includes three modifications to $\exactpn$. 
First, each $n$-gram in the reference can be matched at most once. 
Second, 
the number of exact matches is accumulated for all reference-candidate pairs in the corpus and divided by the total number of $n$-grams in all candidate sentences. 
Finally, very short candidates are discouraged using  a brevity penalty. 
Typically, \bleu{} is computed for multiple values of $n$ (e.g. $n=1,2,3,4$) and the scores are averaged geometrically. 
A smoothed variant, \metric{SentBleu}~\citep{moses} is computed at the sentence level. 
In contrast to \bleu, \method is not restricted to maximum $n$-gram length, but instead relies on contextualized embeddings that are able to capture dependencies of potentially unbounded length.

\paragraph{\meteor}

\meteor~\citep{meteor} computes $\exactp_1$ and $\exactr_1$ while allowing backing-off from  exact unigram matching to matching word stems, synonyms, and paraphrases. 
For example, \nlstring{running} may match \nlstring{run} if no exact match is possible.
Non-exact matching uses an external stemmer, a synonym lexicon, and a paraphrase table. 
\meteor 1.5~\citep{meteor1.5} weighs content and function words differently, and also applies importance weighting to different matching types.
The more recent \metric{Meteor++ 2.0}~\citep{meteor++2.0} further incorporates a learned external paraphrase resource.
Because \meteor requires external resources, only five languages are supported with the full feature set, and eleven are partially supported. 
Similar to \meteor, \method allows relaxed matches, but relies on \bert embeddings that are trained on large amounts of raw text and are currently available for 104 languages. 
\method also supports importance weighting, which we estimate with simple corpus statistics. 

\paragraph{Other Related Metrics}

\metric{NIST}~\citep{nist} is a revised version of \bleu that weighs each $n$-gram differently and uses an alternative brevity penalty.
$\Delta$\bleu~\citep{deltaBLEU} modifies multi-reference \bleu by including human annotated negative reference sentences.
\metric{chrF}~\citep{chrF} compares character $n$-grams in the reference and candidate sentences. \metric{chrF++}~\citep{chrF++} extends \metric{chrF} to include word bigram matching. 
\rouge~\citep{rouge} is a commonly used metric for summarization evaluation.
\rouge-$n$~\citep{rouge} computes $\exactrn$ (usually $n=1,2$), while \rouge-$L$ is a  variant of $\exactr_1$ with the numerator replaced by the length of the longest common subsequence.
\metric{CIDEr}~\citep{cider} is an image captioning metric that computes cosine similarity between $\tf$--$\idf$ weighted $n$-grams. 
We adopt a similar approach to weigh tokens differently. 
Finally, \citet{debiasing} and \citet{unifying} combine automatic metrics with human judgments for text generation evaluation.

\subsection{Edit-distance-based Metrics} 
Several methods use word edit distance or word error rate~\citep{wer}, which quantify similarity using the number of edit operations required to get from the candidate to the reference. 
TER~\citep{ter} normalizes edit distance by the number of reference words, and ITER~\citep{iter} adds stem matching and better normalization. 
\metric{PER}~\citep{per} computes position independent error rate, CDER~\citep{cder} models block reordering as an edit operation.
\metric{CharacTer}~\citep{wang2016character} and \metric{EED}~\citep{eed} operate on the character level and achieve higher correlation with human judgements on some languages.

\subsection{Embedding-based Metrics}
Word embeddings~\citep{word2vec, glove, fasttext, dai2017mixture, athiwaratkun2018probabilistic} are learned dense token representations. 
\meant 2.0~\citep{meant2} uses word embeddings and shallow semantic parses to compute lexical and structural similarity. 
\metric{Yisi-1}~\citep{Lo2018:yisi} is similar  to \meant 2.0, but makes the use of semantic parses optional. 
Both methods use a relatively simple similarity computation, which inspires our approach, including using greedy matching~\citep{Corley2005:greedymatch} and experimenting with a similar importance weighting to \metric{Yisi-1}. 
However, we use contextual embeddings, which capture the specific use of a token in a sentence, and potentially capture sequence information. We do not use external tools to generate linguistic structures, which makes our approach relatively simple and portable to new languages. 
Instead of greedy matching, WMD~\citep{kusner2015word}, WMD\textsubscript{O}~\citep{wmdo}, and SMS~\citep{sms} propose to use optimal matching based on earth mover's distance~\citep{emd}.
% Instead of greedy matching, WMD~\citep{kusner2015word} and SMS~\citep{sms} propose to use optimal matching based on earth mover's distance~\citep{emd}.
The tradeoff\footnote{We provide an ablation study of this design choice in Appendix~\ref{sec:sup-mover-ablate}.} between greedy and optimal matching was studied by \citet{rus2012comparison}. 
\citet{task-dialog-eval} compute similarity with sentence-level representations.  
In contrast, our token-level computation allows us to weigh tokens differently according to their importance.

\subsection{Learned Metrics}
Various metrics are trained to optimize correlation with human judgments. 
\beer~\citep{beer} uses a regression model based on character $n$-grams and word bigrams. 
\blend~\citep{blend} uses regression to combine 29 existing metrics. 
\ruse~\citep{ruse} combines three pre-trained sentence embedding models. 
All these methods require costly human judgments as supervision for each dataset, and risk poor generalization to new domains, even within a known language and task~\citep{debiasing}. 
\citet{leic} and \citet{lowe-turingtest} train a neural model to predict if the input text is human-generated. 
This approach also has the risk of being optimized to existing data and generalizing poorly to new data. In contrast, the model underlying \method is not optimized for any specific evaluation task.

\section{\method}
\label{sec:method}

\begin{figure*}[t]
    \centering
    \includegraphics[width=\linewidth]{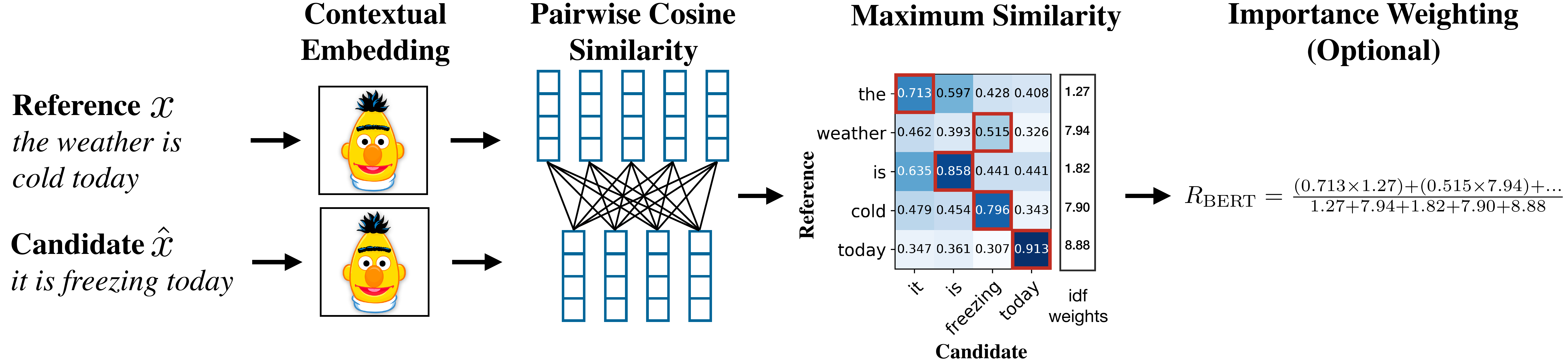}
    \vspace{-20pt}
    \caption{Illustration of the computation of the recall metric $\methodr{}$. Given the reference $\sentref$ and candidate $\senthyp{}$, we compute \bert embeddings and pairwise cosine similarity. We highlight the greedy matching in red, and include the optional $\idf$ importance weighting.}
    \label{fig:bert}
\end{figure*}

Given a reference sentence $\sentref{} = \langle \sentref{}_1, \dots, \sentref{}_k  \rangle$ and a candidate sentence $\senthyp = \langle \senthyp_1, \dots, \senthyp_l\rangle$, we use contextual embeddings to represent the tokens, and compute  matching using cosine similarity, optionally weighted with inverse document frequency scores. 
Figure~\ref{fig:bert} illustrates the computation.

\newcommand{\vref}{\mathbf{x}}
\newcommand{\vhyp}{\mathbf{\hat{x}}}

\paragraph{Token Representation}

We use contextual embeddings to represent the tokens in the input sentences $\sentref{}$ and $\senthyp{}$. 
In contrast to prior word embeddings~\citep{word2vec,glove}, contextual embeddings, such as \bert~\citep{bert} and \textsc{ELMo}~\citep{elmo}, can generate different vector representations for the same word in different sentences depending on the surrounding words, which form the context of the target word. The models used to generate these embeddings are most commonly trained using various language modeling objectives, such as masked word prediction~\citep{bert}. 

We experiment with different models (Section~\ref{sec:exp}), using the tokenizer provided with each model. 
Given a tokenized reference sentence $\sentref{}=\langle \sentref{}_1, \dots, \sentref{}_k  \rangle$, 
the embedding model generates a sequence of vectors $\langle \vref_1, \dots, \vref_k \rangle$. Similarly, the tokenized candidate $\senthyp{} = \langle \senthyp{}_1, \dots, \senthyp{}_m \rangle$ is mapped to $\langle \vhyp_1, \dots, \vhyp_l\rangle$. 
The main model we use is \bert, which tokenizes the input text into a sequence of word pieces~\citep{google16}, where unknown words are split into several commonly observed sequences of characters. 
The representation for each word piece is computed with a  Transformer encoder~\citep{transformer} by  repeatedly applying self-attention and nonlinear transformations in an alternating fashion. 
\bert embeddings have been shown to benefit various NLP tasks~\citep{bert, bert-summarization, bert-emotion, bert-retrieval}. 

\paragraph{Similarity Measure}

The vector representation allows for a soft measure of similarity instead of exact-string~\citep{bleu} or heuristic~\citep{meteor}  matching. 
The cosine similarity of a reference token $\sentref_i$ and a candidate token $\senthyp_j$ is $\frac{\vref_i^\top \vhyp_j}{\|\vref_i\| \|\vhyp_j\|}$. 
We use pre-normalized vectors, which reduces this calculation to the inner product $\vref_i^\top\vhyp_j$.  
While this measure considers tokens in isolation, the contextual embeddings contain information from the rest of the sentence. 

\paragraph{\method} 

The complete score matches each token in $\sentref{}$ to a token in $\senthyp{}$ to compute recall, and each token in $\senthyp{}$ to a token in $\sentref{}$ to compute precision. We use greedy matching to maximize the matching similarity score,\footnote{We compare greedy matching with optimal assignment in Appendix~\ref{sec:sup-mover-ablate}.} where each token is matched to the most similar token in the other sentence. We combine precision and recall to compute an F1 measure. 
For a reference $\sentref$ and candidate $\senthyp$, the recall, precision, and F1 scores are:
\begin{equation*}
    \methodr =\frac{1}{|\sentref|} \sum_{\sentref_i \in \sentref}   \max_{\senthyp_j \in \senthyp{}} \vref_i^\top \vhyp_{j} \nonumber\;\;, \quad
\methodp = \frac{1}{|\senthyp|}  \sum_{\senthyp_j\in \senthyp}   \max_{\sentref_i\in \sentref}  \vref_{i}^\top \vhyp_{j} \nonumber\;\;, \quad 
\methodf = 2\frac{\methodp \cdot \methodr }{ \methodp + \methodr }\;\;.
\end{equation*}

\paragraph{Importance Weighting}

Previous work on similarity measures demonstrated that rare words can be more indicative for sentence similarity than common words~\citep{meteor,cider}. 
\method enables us to easily incorporate importance weighting. 
We experiment with inverse document frequency ($\idf$) scores computed from the test corpus. 
Given $M$ reference sentences $\{\sentref{}^{(i)}\}_{i=1}^M$, the $\idf$ score of a word-piece token $w$ is
\begin{equation*}
\idf(w) = -\log \frac{1}{M}\sum_{i=1}^M \mathbb{I} [w \in \sentref{}^{(i)}] \;\;,
\end{equation*}
where $\mathbb{I}[\cdot]$ is an indicator function. 
We do not use the full $\tf$-$\idf$ measure because we process single sentences, where the term frequency (${\rm tf}$) is likely  1. For example, recall with $\idf$ weighting is
\begin{equation*}
    \methodr = \frac{\sum_{\sentref_i \in \sentref} \idf(\sentref_i)  \max_{\senthyp_j \in \senthyp{}} \vref_i^\top \vhyp_{j}}{\sum_{\sentref_i\in \sentref} \idf(\sentref_i)}\;\;. \nonumber
\end{equation*}
Because we use reference sentences to compute $\idf$, the $\idf$ scores remain the same for all systems evaluated on a specific test set. We apply plus-one smoothing to handle unknown word pieces. 

\paragraph{Baseline Rescaling}

Because we use pre-normalized vectors, our computed scores have the same numerical range of cosine similarity (between $-1$ and $1$).
However, in practice we observe scores in a more limited range, potentially because of  the learned geometry of contextual embeddings. 
While this characteristic does not impact \method's capability to rank text generation systems, it makes the actual score less readable.
We address this by rescaling \method with respect to its empirical lower bound $b$ as a baseline. 
We compute $b$ using Common Crawl monolingual datasets.\footnote{\url{https://commoncrawl.org/}}
For each language and contextual embedding model, we create $1$M candidate-reference pairs by grouping two random sentences. 
Because of the random pairing and the corpus diversity,  each pair has very low lexical and semantic overlapping.\footnote{\bleu computed on these pairs is around zero.} 
We compute $b$ by averaging \method computed on these sentence pairs.
Equipped with baseline $b$, we rescale \method  linearly. For example, the rescaled value $\hatmethodr$ of  $\methodr$ is: 
\begin{equation*}
    \hatmethodr = \frac{\methodr-b}{1-b}\;\;. \nonumber
\end{equation*}
After this operation $\hatmethodr$ is typically between $0$ and $1$. 
We apply the same rescaling procedure for $\methodp$ and $\methodf$.
This method does not affect the ranking ability and human correlation of \method, and is intended solely to increase the score readability.

\section{Experimental Setup}
\label{sec:exp}

We evaluate our approach on machine translation and image captioning.

\paragraph{Contextual Embedding Models}
We evaluate twelve pre-trained contextual embedding models, including variants of BERT~\citep{bert}, RoBERTa~\citep{roberta}, XLNet~\citep{yang2019xlnet}, and XLM~\citep{xlm}. 
We present the best-performing models in Section~\ref{sec:results}. 
We use the 24-layer $\roberta$ model\footnote{We use the tokenizer provided with each model. For all Hugging Face models that use the GPT-2 tokenizer, at the time of our experiments, the tokenizer adds a space to the beginning of each sentence.} for English tasks, 12-layer $\bertchinese$ model for Chinese tasks, and the 12-layer cased multilingual $\bertmulti$ model for other languages.\footnote{All the models used are from  \href{https://github.com/huggingface/pytorch-transformers}{https://github.com/huggingface/pytorch-transformers}.} 
We show the performance of all other models in Appendix~\ref{sec:sup-additiona-data}.
Contextual embedding models generate embedding representations at every layer in the encoder network. 
Past work has shown that intermediate layers produce more effective representations for semantic tasks~\citep{linguistic-context}. 
We use the WMT16 dataset ~\citep{wmt16em} as a validation set to select the best layer of each model (Appendix~\ref{sec:bert-study}).

\paragraph{Machine Translation}
Our main evaluation corpus is the WMT18 metric evaluation dataset~\citep{wmt18em}, which contains predictions of 149 translation systems across 14 language pairs, gold references, and two types of human judgment scores. 
Segment-level human judgments assign a score to each reference-candidate pair. System-level human judgments associate each system with a single score based on all pairs in the test set.
WMT18 includes translations from English to Czech, German, Estonian, Finnish, Russian, and Turkish, and from the same set of languages to English.
We follow the WMT18 standard practice and use absolute Pearson correlation $\lvert \rho \rvert$ and Kendall rank correlation $\tau$  to evaluate metric quality, and compute significance with the Williams test~\citep{williams1959regression} for $\lvert \rho \rvert$ and bootstrap re-sampling for $\tau$ as suggested by \citet{graham2014testing}.
We compute system-level scores by averaging \method{} for every reference-candidate pair. We also experiment with hybrid systems by randomly sampling one candidate sentence from one of the available systems for each reference sentence~\citep{super-sample}. This enables system-level experiments with a higher number of systems. Human judgments of each hybrid system are created by averaging the WMT18 segment-level human judgments for the corresponding sentences in the sampled data.
We compare \method{}s to one canonical metric for each category introduced in Section~\ref{sec:related_works}, and include the comparison with all other participating metrics from WMT18 in Appendix~\ref{sec:sup-additiona-data}.

In addition to the standard evaluation, we design model selection experiments. We use 10K hybrid systems super-sampled from WMT18. 
We randomly select 100 out of 10K hybrid systems, and rank them using the automatic metrics. We repeat this process 100K times. 
We report the percentage of the metric ranking agreeing with the human ranking on the best system (Hits@1). 
In Tables~\ref{tab:sup-wmt18-to-modelselect-hit1}-\ref{tab:sup-wmt18-from-modelselect-diff}, we include two additional measures to the model selection study: (a) the mean reciprocal rank of the top metric-rated system according to the human ranking,  and (b) the difference between the human score of the top human-rated system and that of the top metric-rated system.

Additionally, we report the same study on the WMT17~\citep{wmt17em} and the WMT16~\citep{wmt16em} datasests in Appendix~\ref{sec:sup-additiona-data}.\footnote{For WMT16, we only conduct segment-level experiments on to-English pairs due to errors in the dataset.} 
This adds 202 systems to our evaluation. 

\paragraph{Image Captioning}

We use the human judgments of twelve submission entries from the COCO 2015 Captioning  Challenge. 
Each participating system generates a caption for each image in the COCO validation set~\citep{coco}, and each image has approximately five reference captions.
Following \citet{leic}, we compute the Pearson correlation with two  system-level  metrics: the percentage of captions that are evaluated as better or equal to human captions (M1) and the percentage of captions that are indistinguishable from human captions (M2). We compute \method with multiple references by scoring the candidate with each available reference and returning the highest score. 
We compare with eight task-agnostic metrics: \bleu~\citep{bleu}, \meteor ~\citep{meteor}, \rouge-L~\citep{rouge}, \metric{CIDEr}~\citep{cider}, \metric{BEER}~\citep{beer},
EED~\citep{eed}, \metric{chrF++}~\citep{chrF++}, and \metric{CharacTER}~\citep{wang2016character}. We also compare with two task-specific metrics:  \spice~\citep{spice} and \leic~\citep{leic}. 
\spice is computed using the similarity of scene graphs parsed from the reference and candidate captions.
\leic is trained to predict if a caption is written by a human given the image. 

\begin{table*}[t!]
    \footnotesize
    \centering
    \begin{tabular}{cccccccc}
\toprule
Metric & en$\leftrightarrow$cs & en$\leftrightarrow$de & en$\leftrightarrow$et & en$\leftrightarrow$fi & en$\leftrightarrow$ru & en$\leftrightarrow$tr & en$\leftrightarrow$zh\\
 & (5/5) & (16/16) & (14/14) & (9/12) & (8/9) & (5/8) & (14/14)\\
\midrule
BLEU & .970/\textbf{.995} & .971/\textbf{.981} & \textbf{.986}/\textbf{.975} & .973/\textbf{.962} & .979/\textbf{.983} & \textbf{.657}/.826 & .978/.947\\
ITER & .975/.915 & .990/\textbf{.984} & .975/\textbf{.981} & \textbf{.996}/\textbf{.973} & .937/.975 & \textbf{.861}/.865 & .980/\phantom{.0}--\phantom{0}\\
RUSE & .981/\phantom{.0}--\phantom{0} & .997/\phantom{.0}--\phantom{0} & \textbf{.990}/\phantom{.0}--\phantom{0} & .991/\phantom{.0}--\phantom{0} & \textbf{.988}/\phantom{.0}--\phantom{0} & \textbf{.853}/\phantom{.0}--\phantom{0} & \textbf{.981}/\phantom{.0}--\phantom{0}\\
YiSi-1 & .950/\textbf{.987} & .992/\textbf{.985} & .979/\textbf{.979} & .973/.940 & \textbf{.991}/\textbf{.992} & \textbf{.958}/\textbf{.976} & .951/\textbf{.963}\\
$P_{\text{BERT}}$ & .980/\textbf{.994} & \textbf{.998}/\textbf{.988} & \textbf{.990}/\textbf{.981} & .995/.957 & .982/\textbf{.990} & \textbf{.791}/\textbf{.935} & .981/.954\\
$R_{\text{BERT}}$ & \textbf{.998}/\textbf{.997} & .997/\textbf{.990} & .986/\textbf{.980} & \textbf{.997}/\textbf{.980} & \textbf{.995}/\textbf{.989} & .054/.879 & \textbf{.990}/\textbf{.976}\\
$F_{\text{BERT}}$ & \textbf{.990}/\textbf{.997} & \textbf{.999}/\textbf{.989} & .990/\textbf{.982} & \textbf{.998}/\textbf{.972} & \textbf{.990}/.990 & \textbf{.499}/.908 & \textbf{.988}/.967\\
$F_{\text{BERT}}$ (idf) & .985/\textbf{.995} & \textbf{.999}/\textbf{.990} & \textbf{.992}/\textbf{.981} & .992/\textbf{.972} & \textbf{.991}/\textbf{.991} & \textbf{.826}/\textbf{.941} & \textbf{.989}/\textbf{.973}\\
\bottomrule
\end{tabular}
    \vspace{-2pt}
    \caption{Absolute Pearson correlations with system-level human judgments on WMT18. For each language pair, the left number is the to-English correlation, and the right is the from-English. We bold correlations of metrics not significantly outperformed  by any other metric under Williams Test for that language pair and direction. The numbers in parenthesis are the number of systems used for each language pair and direction.}
    \label{tab:wmt18-sys}
\end{table*}

\begin{table*}[t!]
    \footnotesize
    \centering
    \begin{tabular}{cccccccc}
\toprule
Metric & en$\leftrightarrow$cs & en$\leftrightarrow$de & en$\leftrightarrow$et & en$\leftrightarrow$fi & en$\leftrightarrow$ru & en$\leftrightarrow$tr & en$\leftrightarrow$zh\\
\midrule
BLEU & .956/.993 & .969/\textbf{.977} & \textbf{.981}/.971 & .962/.958 & .972/.977 & .586/.796 & .968/.941\\
ITER & .966/.865 & .990/.978 & .975/\textbf{.982} & .989/.966 & .943/.965 & .742/.872 & .978/\phantom{.0}--\phantom{0}\\
RUSE & .974/\phantom{.0}--\phantom{0} & .996/\phantom{.0}--\phantom{0} & .988/\phantom{.0}--\phantom{0} & \textbf{.983}/\phantom{.0}--\phantom{0} & .982/\phantom{.0}--\phantom{0} & .780/\phantom{.0}--\phantom{0} & .973/\phantom{.0}--\phantom{0}\\
YiSi-1 & .942/.985 & .991/.983 & .976/.976 & .964/.938 & \textbf{.985}/\textbf{.989} & \textbf{.881}/\textbf{.942} & .943/.957\\
$P_{\text{BERT}}$ & .965/.989 & .995/.983 & \textbf{.990}/\textbf{.970} & .976/.951 & .976/.988 & .846/.936 & .975/.950\\
$R_{\text{BERT}}$ & \textbf{.989}/\textbf{.995} & .997/\textbf{.991} & .982/\textbf{.979} & .989/\textbf{.977} & \textbf{.988}/\textbf{.989} & .540/\textbf{.872} & \textbf{.981}/\textbf{.980}\\
$F_{\text{BERT}}$ & .978/\textbf{.993} & .998/.988 & .989/.978 & .983/.969 & .985/.989 & .760/.910 & \textbf{.981}/.969\\
$F_{\text{BERT}}$ (idf) & .982/.995 & \textbf{.998}/.988 & \textbf{.988}/.979 & \textbf{.989}/.969 & .983/.987 & .453/.877 & .980/.963\\
\bottomrule
\end{tabular}
    \vspace{-2pt}
    \caption{Absolute Pearson correlations with system-level human judgments on WMT18. We use  10K hybrid super-sampled systems for each language pair and direction. For each language pair, the left number is the to-English correlation, and the right is the from-English.
    Bolding criteria is the same as in Table~\ref{tab:wmt18-sys}. }
    \label{tab:wmt18-sys-hybrids}
\end{table*}

\begin{table*}[t!]
    \footnotesize
    \centering
    \begin{tabular}{cccccccc}
\toprule
Metric & en$\leftrightarrow$cs & en$\leftrightarrow$de & en$\leftrightarrow$et & en$\leftrightarrow$fi & en$\leftrightarrow$ru & en$\leftrightarrow$tr & en$\leftrightarrow$zh\\
\midrule
BLEU & .134/.151 & .803/.610 & .756/.618 & .461/.088 & .228/.519 & .095/.029 & .658/.515\\
ITER & .154/.000 & .814/.692 & .742/.733 & .475/.111 & .234/.532 & .102/.030 & .673/\phantom{.0}--\phantom{0}\\
RUSE & \textbf{.214}/\phantom{.0}--\phantom{0} & .823/\phantom{.0}--\phantom{0} & \textbf{.785}/\phantom{.0}--\phantom{0} & .487/\phantom{.0}--\phantom{0} & .248/\phantom{.0}--\phantom{0} & .109/\phantom{.0}--\phantom{0} & .670/\phantom{.0}--\phantom{0}\\
YiSi-1 & .159/.178 & .809/.671 & .749/.671 & .467/\textbf{.230} & .248/.544 & .108/\textbf{.398} & .613/.594\\
$P_{\text{BERT}}$ & .173/.180 & .706/.663 & .764/\textbf{.771} & .498/.078 & .255/\textbf{.545} & .140/.372 & .661/.551\\
$R_{\text{BERT}}$ & .163/\textbf{.184} & .804/\textbf{.730} & .770/.722 & .494/.148 & .260/.542 & .005/.030 & .677/\textbf{.657}\\
$F_{\text{BERT}}$ & .175/.184 & \textbf{.824}/.703 & .769/.763 & .501/.082 & .262/.544 & \textbf{.142}/.031 & .673/.629\\
$F_{\text{BERT}}$ (idf) & .179/.178 & \textbf{.824}/.722 & .760/.764 & \textbf{.503}/.082 & \textbf{.265}/.539 & .004/.030 & \textbf{.678}/.595\\
\bottomrule
\end{tabular}
    \vspace{-2pt}
    \caption{Model selection accuracies (Hits@1) on WMT18 hybrid systems. We report the average of 100K samples and the 0.95 confidence intervals are below $10^{-3}$. We bold the highest numbers for each language pair and direction.}
    \label{tab:wmt18-model-select}
\end{table*}

\begin{table*}[t!]
    \footnotesize
    \centering
    
\begin{tabular}{cccccccc}
\toprule
Metric & en$\leftrightarrow$cs & en$\leftrightarrow$de & en$\leftrightarrow$et & en$\leftrightarrow$fi & en$\leftrightarrow$ru & en$\leftrightarrow$tr & en$\leftrightarrow$zh\\
 & (5k/5k) & (78k/ 20k) & (57k/32k) & (16k/10k) & (10k/22k) & (9k/1k) & (33k/29k)\\
\midrule
BLEU & .233/.389 & .415/.620 & .285/.414 & .154/.355 & .228/.330 & .145/.261 & .178/.311 \\
ITER & .198/.333 & .396/.610 & .235/.392 & .128/.311 & .139/.291 & -.029/.236 & .144/\phantom{.0}--\phantom{0} \\
RUSE & .347/\phantom{.0}--\phantom{0} & .498/\phantom{.0}--\phantom{0} & .368/\phantom{.0}--\phantom{0} & .273/\phantom{.0}--\phantom{0} & .311/\phantom{.0}--\phantom{0} & .259/\phantom{.0}--\phantom{0} & .218/\phantom{.0}--\phantom{0}\\
YiSi-1 & .319/.496 & .488/.691 & .351/.546 & .231/.504 & .300/.407 & .234/.418 & .211/.323 \\
$P_{\text{BERT}}$ & .387/.541 & .541/.715 & .389/.549 & .283/.486 & .345/.414 & .280/.328 & .248/.337 \\
$R_{\text{BERT}}$ & .388/\textbf{.570} & .546/\textbf{.728} & .391/\textbf{.594} & \textbf{.304}/\textbf{.565} & .343/.420 & .290/\textbf{.411} & .255/\textbf{.367} \\
$F_{\text{BERT}}$ & .404/.562 & \textbf{.550}/\textbf{.728} & \textbf{.397}/.586 & .296/.546 & \textbf{.353}/.423 & .292/.399 & \textbf{.264}/.364 \\
$F_{\text{BERT}}$ (idf) & \textbf{.408}/.553 & \textbf{.550}/.721 & .395/585 & .293/.537 & .346/\textbf{.425} & \textbf{.296}/.406 & .260/.366 \\
\bottomrule
\end{tabular}
    \vspace{-2pt}
    \caption{Kendall correlations with segment-level human judgments on WMT18. For each language pair, the left number is the to-English correlation, and the right is the from-English. We bold correlations of metrics not significantly outperformed by any other metric under bootstrap sampling for that language pair and direction. The numbers in parenthesis are the number of candidate-reference sentence pairs for each language pair and direction.}
    \label{tab:wmt18-seg}
\end{table*}

\section{Results}
\label{sec:results}

\paragraph{Machine Translation}
Tables~\ref{tab:wmt18-sys}--\ref{tab:wmt18-model-select} show system-level correlation to human judgements, correlations on hybrid systems, and model selection performance. 
We observe that \method is consistently a top performer. 
In to-English results, RUSE~\citep{ruse} shows competitive performance. However, RUSE is a supervised method trained on WMT16 and WMT15 human judgment data. In cases where RUSE models were not made available, such as for our from-English experiments, it is not possible to use RUSE without additional data and training. 
Table~\ref{tab:wmt18-seg} shows segment-level correlations. We see that \method exhibits significantly higher performance compared to the other metrics. 
The large improvement over \bleu stands out, making \method particularly suitable to analyze specific examples, where \sentbleu is less reliable. 
In Appendix~\ref{sec:sup-qualitative}, we provide qualitative examples to illustrate the segment-level performance difference between \sentbleu and \method.
At the segment-level, \method even significantly outperforms RUSE. 
Overall, we find that applying importance weighting using $\idf$ at times provides small benefit, but in other cases does not help. 
Understanding better when such importance weighting is likely to help is an important direction for future work, and likely depends on the domain of the text and the available test data. 
We continue without $\idf$ weighting for the rest of our experiments. 
While recall $\methodr$, precision $\methodp$, and F1 $\methodf$  alternate as the best measure in different setting, F1 $\methodf$ performs reliably well across all the different settings.
Our overall recommendation is therefore to use F1. 
We present additional results using the full set of 351 systems and evaluation metrics in Tables~\ref{tab:sup-wmt16-to-seg}--\ref{tab:sup-wmt18-from-modelselect-diff} in the appendix, including for experiments with $\idf$ importance weighting,  different contextual embedding models, and model selection. 

\paragraph{Image Captioning}

Table~\ref{tab:coco} shows correlation results for the COCO Captioning Challenge. 
\method outperforms all task-agnostic baselines by large margins. 
Image captioning presents a challenging evaluation scenario, and metrics based on strict $n$-gram matching, including \bleu and \metric{Rouge}, show weak correlations with human judgments.
$\idf$ importance weighting shows significant benefit for this task, suggesting people attribute higher importance to content words.
Finally, \leic~\citep{leic}, a trained metric  that takes images as additional inputs and is optimized specifically for the COCO data and this set of systems, outperforms all other methods. 

\begin{table}[t]
    \footnotesize
    \centering
    \begin{minipage}[m]{0.40\linewidth}
    \centering
    \vspace{-0.12in}
    \resizebox{\linewidth}{!}{
    \begin{tabular}{ccc}
    \toprule
    Metric & M1 & M2 \\
    \midrule
    \bleu & -0.019$^*$ & -0.005$^*$ \\
    \meteor & 0.606$^*$ & 0.594$^*$ \\
    \rouge-L & 0.090$^*$ & 0.096$^*$ \\
    \metric{CIDEr} & 0.438$^*$ & 0.440$^*$ \\
    \spice & 0.759$^*$ & 0.750$^*$ \\
    \leic & \textbf{0.939}$^*$ & \textbf{0.949}$^*$ \\
    \metric{BEER} & 0.491 & 0.562 \\
    \metric{EED} & 0.545 & 0.599 \\
    \metric{chrF++} & 0.702 & 0.729 \\
    \metric{CharacTER} & 0.800 & 0.801 \\
    \midrule
    $P_{\text{BERT}}$ & -0.105 & -0.041 \\
    $R_{\text{BERT}}$ & 0.888 & 0.863 \\
    $F_{\text{BERT}}$ & 0.322 & 0.350 \\
    $R_{\text{BERT}}$ (idf) & \textbf{0.917} & \textbf{0.889} \\
    \bottomrule
\end{tabular}
    }
    \caption{Pearson correlation on the 2015 COCO Captioning Challenge. 
    The M1 and M2 measures are described in Section~\ref{sec:exp}. \leic uses images as additional inputs.
    Numbers with $^*$ are cited from \citet{leic}.
    We bold the highest correlations of task-specific and task-agnostic metrics.
    }
    \label{tab:coco}
    \end{minipage}\quad%
    \begin{minipage}[m]{0.55\linewidth}
    \centering
    \resizebox{\linewidth}{!}{
    \begin{tabular}{c|ccc}
    \toprule
    Type & Method & QQP & PAWS\textsubscript{QQP} \\
    \midrule
    \multirow{3}{*}{ \shortstack[l]{Trained on QQP \\ (supervised)} } & DecAtt & 0.939$^*$ & 0.263 \\
    & DIIN & 0.952$^*$ & 0.324 \\
    & BERT & \textbf{0.963}$^*$ & \textbf{0.351} \\
    \midrule
    \multirow{3}{*}{ \shortstack[l]{Trained on QQP \\+ PAWS\textsubscript{QQP} \\ (supervised)}} & DecAtt & - & 0.511 \\
    & DIIN & - & 0.778 \\
    & BERT & - & \textbf{0.831} \\
    \midrule
    \multirow{11}{*}{ \shortstack[l]{Metric\\(Not trained\\on QQP or \\PAWS\textsubscript{QQP})}} 
    & \bleu & 0.707 & 0.527 \\
    & \metric{Meteor} & 0.755 & 0.532 \\
    & \metric{Rouge-L} & 0.740 & 0.536 \\
    & \metric{chrF++} & 0.577 & 0.608 \\
    & \metric{BEER} & 0.741 & 0.564 \\
    & \metric{EED} & 0.743 & 0.611 \\
    & \metric{CharacTER} &	0.698 & 0.650 \\
    \cmidrule{2-4}
    & $P_{\text{BERT}}$ & 0.757 & 0.687 \\
    & $R_{\text{BERT}}$ & 0.744 & 0.685 \\
    & $F_{\text{BERT}}$ & 0.761 & 0.685 \\
    & $F_{\text{BERT}}$ (idf) & \textbf{0.777} & \textbf{0.693} \\
    \bottomrule
\end{tabular}
    }
    \caption{Area under ROC curve (AUC) on QQP and PAWS\textsubscript{QQP} datasets. The scores of trained DecATT~\citep{decatt}, DIIN~\citep{diin}, and fine-tuned BERT are reported by \citet{paws}. Numbers with $^*$ are  scores on the held-out test set of QQP.
    We bold the highest correlations of task-specific and task-agnostic metrics.
    }
    \label{tab:paws}
    \end{minipage}
    
    \vspace{-10pt}
\end{table}

\paragraph{Speed}

Despite the use of a large pre-trained model, computing \method is relatively fast. We are able to process 192.5 candidate-reference pairs/second using a GTX-1080Ti GPU. 
The complete WMT18 en-de test set, which includes 2,998 sentences,  takes 15.6sec to process, compared to 5.4sec with SacreBLEU~\citep{sacrebleu}, a common \bleu implementation. Given the sizes of commonly used test and validation sets, the increase in processing time is relatively marginal, and \method is a good fit for using during validation (e.g., for stopping) and testing, especially when compared to the time costs of other development stages. 

\section{Robustness Analysis}
\label{sec:robust}

We test the robustness of \method using adversarial paraphrase classification. 
We use the Quora Question Pair corpus~\citep[QQP;][]{QQP} and the adversarial paraphrases from the Paraphrase Adversaries from Word Scrambling dataset~\citep[PAWS;][]{paws}. 
Both datasets contain pairs of sentences labeled to indicate whether they are paraphrases or not. 
Positive examples in QQP are real duplicate questions, while negative examples are related, but different questions. 
Sentence pairs in PAWS are generated through word swapping. 
For example, in PAWS, \nlstring{Flights from New York to Florida} may be changed to \nlstring{Flights from Florida to New York} and a good classifier should identify that these two sentences are not paraphrases.
PAWS includes two parts: PAWS\textsubscript{QQP}, which is based on the QQP data, and PAWS\textsubscript{Wiki}. 
We use the PAWS\textsubscript{QQP} development set which contains 667 sentences. 
% extracted from QQP dataset. 
For the automatic metrics, we use no paraphrase detection training data. 
We expect that pairs with higher scores are more likely to be paraphrases.
To evaluate the automatic metrics on QQA, we use the first 5,000 sentences in the training set instead of the the test set because the test labels are not available. 
We treat the first sentence as the reference and the second sentence as the candidate.

Table~\ref{tab:paws} reports the area under ROC curve (AUC) for  existing models and automatic metrics. 
We observe that supervised classifiers trained on QQP perform worse than random guess on PAWS\textsubscript{QQP}, which shows these models predict the adversarial examples are more likely to be paraphrases. 
When adversarial examples are provided in training, state-of-the-art models like DIIN~\citep{diin} and fine-tuned BERT are able to identify the adversarial examples but their performance still decreases significantly from their performance on QQP.
Most metrics have decent performance on QQP, but show a significant performance drop on  PAWS\textsubscript{QQP}, almost down to chance performance. This suggests these metrics fail to to distinguish the harder adversarial examples. 
In contrast, the performance of \method drops only slightly, showing  more robustness than the other metrics.

\section{Discussion}
\label{sec:discussion}

We propose \method, a new metric for evaluating generated text against gold standard references. 
\method is purposely designed to be simple, task agnostic, and easy to use. 
Our analysis illustrates how \method resolves some of the limitations of commonly used metrics, especially on challenging adversarial examples.
We conduct extensive experiments with various configuration choices for \method, including the contextual embedding model used and the use of importance weighting. 
Overall, our extensive experiments, including the ones in the appendix, show that \method  achieves better correlation than common metrics, and is effective for model selection. 
However, there is no one configuration of \method that clearly outperforms all others. 
While the differences between the top configurations are often small, it is important for the user to be aware of the different trade-offs, and consider the domain and languages when selecting the exact configuration to use. 
In general, for machine translation evaluation, we suggest using $\methodf$, which we find the most reliable. 
For evaluating text generation in English, we recommend using the 24-layer $\roberta$ model to compute $\method$.
For non-English language, the multilingual $\bertmulti$ is a suitable choice although $\method$ computed with this model has less stable performance on low-resource languages.
We report the optimal hyperparameter for all models we experimented with in Appendix~\ref{sec:bert-study}

Briefly following our initial preprint publication, \citet{moverscore} published a concurrently developed method related to ours, but with a focus on integrating contextual word embeddings with earth mover's distance~\citep[EMD;][]{emd} rather than our simple matching process. They also propose various improvements compared to our use of contextualized embeddings. 
We study these improvements in Appendix~\ref{sec:sup-mover-ablate} and show that integrating them into \method makes it equivalent or better than the EMD-based approach. Largely though, the effect of the different improvements on \method is more modest compared to their method. 
Shortly after our initial publication, YiSi-1 was updated to use BERT embeddings, showing improved performance~\citep{lo-2019-yisi}. This further corroborates our findings. 
Other recent related work includes training a model on top of BERT to maximize the correlation with human judgments~\citep{bertesim} and evaluating generation with a BERT model fine-tuned on paraphrasing~\citep{yoshimura-etal-2019-filtering}. 
More recent work shows the potential of using \method for training a summarization system~\citep{li-etal-2019-deep} and for domain-specific evaluation using SciBERT~\citep{scibert} to evaluate  abstractive text summarization~\citep{abstract-gen}.

In future work,  we look forward to designing new task-specific metrics that use \method as a subroutine and accommodate task-specific needs, similar to how \citet{beyond-bleu} suggests to use semantic similarity for machine translation training. 
Because \method{} is fully differentiable, it also can be incorporated into a training procedure to compute a learning loss that reduces the mismatch between optimization and evaluation objectives.

\section*{Acknowledgement}
This research is supported in part by grants from the National
Science Foundation (III-1618134, III-1526012, IIS1149882,
IIS-1724282, TRIPODS-1740822, CAREER-1750499), the Office
of Naval Research DOD (N00014-17-1-2175), and the
Bill and Melinda Gates Foundation, SAP, Zillow, Workday, and Facebook Research. 
We thank Graham Neubig and David Grangier for  for their insightful comments. We thank the Cornell NLP community including but not limited to Claire Cardie, Tianze Shi, Alexandra Schofield, Gregory Yauney, and Rishi Bommasani. 
We thank Yin Cui and Guandao Yang for their help with the COCO 2015 dataset.

\bibliography{ref}
\bibliographystyle{iclr2020_conference}

\newpage
\appendix

\section{Qualitative Analysis}
\label{sec:sup-qualitative}

\begin{table*}[t]
    \centering
    \footnotesize
    \resizebox{\textwidth}{!}{%
    \begin{tabular}{l|c|l|c|c|c}
    \toprule
         Case& No. & Reference and Candidate Pairs & Human & $\methodf{}$ & \bleu \\ 
        \midrule
        \parbox[t]{2mm}{\multirow{10}{*}{\rotatebox[origin=c]{90}{$\methodf{}$  $>$ \bleu}}}
        
        & \multirow{2}{*}{1.} & $\sentref$: At the same time Kingfisher is closing 60 B\&Q outlets across the country  & \multirow{2}{*}{38} & \multirow{2}{*}{125} & \multirow{2}{*}{530} \\ 
        & & $\senthyp$: At the same time, Kingfisher will close 60 B \& Q stores nationwide &  &  & \\
        \cline{2-6}
        
        & \multirow{2}{*}{2.} & $\sentref$: Hewlett-Packard to cut up to 30,000 jobs  & \multirow{2}{*}{119} & \multirow{2}{*}{39} & \multirow{2}{*}{441} \\ 
        & & $\senthyp$: Hewlett-Packard will reduce jobs up to 30.000 &  &  & \\
        \cline{2-6}
        
        & \multirow{2}{*}{3.} & $\sentref$: According to opinion in Hungary, Serbia is ``a safe third country".  & \multirow{2}{*}{23} & \multirow{2}{*}{96} & \multirow{2}{*}{465} \\ 
        & & $\senthyp$: According to Hungarian view, Serbia is a ``safe third country." &  &  & \\
        \cline{2-6}

        & \multirow{2}{*}{4.} & $\sentref$: Experts believe November's Black Friday could be holding back spending.  & \multirow{2}{*}{73} & \multirow{2}{*}{147} & \multirow{2}{*}{492} \\ 
        & & $\senthyp$: Experts believe that the Black Friday in November has put the brakes on spending  &  &  & \\
        \cline{2-6}
        
        & \multirow{2}{*}{5.} & $\sentref$: And it's from this perspective that I will watch him die.  & \multirow{2}{*}{37} & \multirow{2}{*}{111} & \multirow{2}{*}{414} \\ 
        & & $\senthyp$: And from this perspective, I will see him die. &  &  & \\
        
        \midrule
        \parbox[t]{2mm}{\multirow{10}{*}{\rotatebox[origin=c]{90}{\bleu $>$ $\methodf{}$}}}  
        & \multirow{2}{*}{6.} & $\sentref$: In their view the human dignity of the man had been violated.  & \multirow{2}{*}{500} & \multirow{2}{*}{470} & \multirow{2}{*}{115} \\ 
        & & $\senthyp$: Look at the human dignity of the man injured. &  &  & \\
        \cline{2-6}
        
        & \multirow{2}{*}{8.} & $\sentref$: For example when he steered a shot from Ideye over the crossbar in the 56th minute.  & \multirow{2}{*}{516} & \multirow{2}{*}{524} & \multirow{2}{*}{185} \\ 
        & & $\senthyp$: So, for example, when he steered a shot of Ideye over the latte (56th). &  &  & \\
        \cline{2-6}
        
        & \multirow{2}{*}{7.} & $\sentref$: A good prank is funny, but takes moments to reverse.  & \multirow{2}{*}{495} & \multirow{2}{*}{424} & \multirow{2}{*}{152} \\ 
        & & $\senthyp$: A good prank is funny, but it takes only moments before he becomes a boomerang. &  &  & \\
        \cline{2-6}

        & \multirow{2}{*}{9.} & $\sentref$: I will put the pressure on them and onus on them to make a decision.   & \multirow{2}{*}{507} & \multirow{2}{*}{471} & \multirow{2}{*}{220} \\ 
        & & $\senthyp$: I will exert the pressure on it and her urge to make a decision. &  &  & \\
        \cline{2-6}
        & \multirow{2}{*}{10.} & $\sentref$: Transport for London is not amused by this flyposting "vandalism."   & \multirow{2}{*}{527} & \multirow{2}{*}{527} & \multirow{2}{*}{246} \\ 
        & & $\senthyp$: Transport for London is the Plaka animal "vandalism" is not funny. &  &  & \\
        \midrule
        
        \parbox[t]{2mm}{\multirow{10}{*}{\rotatebox[origin=c]{90}{$\methodf{}$ $>$ Human}}} 
        & \multirow{2}{*}{11.} & $\sentref$: One big obstacle to access to the jobs market is the lack of knowledge of the German language.  & \multirow{2}{*}{558} & \multirow{2}{*}{131} & \multirow{2}{*}{313} \\ 
        & & $\senthyp$: A major hurdle for access to the labour market are a lack of knowledge of English. &  &  & \\
        \cline{2-6}
        & \multirow{2}{*}{12.} & $\sentref$: On Monday night Hungary closed its 175 km long border with Serbia.  & \multirow{2}{*}{413} & \multirow{2}{*}{135} & \multirow{2}{*}{55} \\ 
        & & $\senthyp$: Hungary had in the night of Tuesday closed its 175 km long border with Serbia. &  &  & \\
        \cline{2-6}
        & \multirow{2}{*}{13.} & $\sentref$: They got nothing, but they were allowed to keep the clothes.  & \multirow{2}{*}{428} & \multirow{2}{*}{174} & \multirow{2}{*}{318} \\ 
        & & $\senthyp$: You got nothing, but could keep the clothes. &  &  & \\
        \cline{2-6}
        & \multirow{2}{*}{14.} & $\sentref$: A majority of Republicans don't see Trump's temperament as a problem.   & \multirow{2}{*}{290} & \multirow{2}{*}{34} & \multirow{2}{*}{134} \\ 
        & & $\senthyp$: A majority of Republicans see Trump's temperament is not a problem. &  &  & \\
        \cline{2-6}
        & \multirow{2}{*}{15.} & $\sentref$:His car was still running in the driveway.  & \multirow{2}{*}{299} & \multirow{2}{*}{49} & \multirow{2}{*}{71} \\ 
        & & $\senthyp$: His car was still in the driveway. &  &  & \\
        \midrule
        \parbox[t]{2mm}{\multirow{10}{*}{\rotatebox[origin=c]{90}{Human $>$ $\methodf{}$}}}  &
         \multirow{2}{*}{16.} & $\sentref$:  Currently the majority of staff are men.  & \multirow{2}{*}{77} & \multirow{2}{*}{525} & \multirow{2}{*}{553} \\ 
        & & $\senthyp$: At the moment the men predominate among the staff. &  &  & \\
        \cline{2-6}
        & \multirow{2}{*}{17.} & $\sentref$: There are, indeed, multiple variables at play.  & \multirow{2}{*}{30} & \multirow{2}{*}{446} & \multirow{2}{*}{552} \\ 
        & & $\senthyp$: In fact, several variables play a role. &  &  & \\
        \cline{2-6}
        & \multirow{2}{*}{18.} & $\sentref$: One was a man of about 5ft 11in tall.  & \multirow{2}{*}{124} & \multirow{2}{*}{551} & \multirow{2}{*}{528} \\ 
        & & $\senthyp$: One of the men was about 1,80 metres in size. &  &  & \\
        \cline{2-6}
        & \multirow{2}{*}{19.} & $\sentref$: All that stuff sure does take a toll.   & \multirow{2}{*}{90} & \multirow{2}{*}{454} & \multirow{2}{*}{547} \\ 
        & & $\senthyp$: All of this certainly exacts its toll. &  &  & \\
        \cline{2-6}
        & \multirow{2}{*}{20.} & $\sentref$: Wage gains have shown signs of picking up.   & \multirow{2}{*}{140} & \multirow{2}{*}{464} & \multirow{2}{*}{514} \\ 
        & & $\senthyp$: Increases of wages showed signs of a recovery. &  &  & \\
    \bottomrule
    \end{tabular}
    }
    \caption{Examples sentences where similarity ranks assigned by Human, $\methodf{}$, and \bleu differ significantly on WMT16 German-to-English evaluation task.   $\sentref$: gold reference, $\senthyp$: candidate outputs of MT systems. Rankings assigned by Human, $\methodf{}$, and \bleu are shown in the right three columns. The sentences are ranked by the similarity, \emph{i.e.} rank 1 is the most similar pair assigned by a score. An ideal metric should rank similar to humans.}
    \label{tab:sup-example-table}
\end{table*}

We study \method and \sentbleu using WMT16 German-to-English~\citep{wmt16em}. 
We rank all 560 candidate-reference pairs by human score, \method, or \sentbleu from most similar to least similar. 
Ideally, the ranking assigned by \method and \sentbleu should be similar to the ranking assigned by the human score. 

Table~\ref{tab:sup-example-table} first shows examples where \method and \sentbleu scores disagree about the ranking for a candidate-reference pair by a large number.
We observe that \method is effectively able to capture synonyms and changes in word order. 
For example, the reference and candidate sentences in pair 3 are almost identical except that the candidate replaces \nlstring{opinion in Hungary} with \nlstring{Hungarian view} and switches the order of the quotation mark (\nlstring{``}) and \nlstring{a}. 
While \method ranks the pair relatively high, \sentbleu judges the pair as dissimilar, because it cannot match synonyms and is sensitive to the small word order changes. 
Pair 5 shows a set of changes that preserve the semantic meaning: replacing \nlstring{to cut} with \nlstring{will reduce} and swapping the order of \nlstring{30,000} and \nlstring{jobs}.
\method ranks the candidate translation similar to the human judgment, whereas \sentbleu ranks it much lower. 
We also see that \sentbleu potentially over-rewards $n$-gram overlap, even when phrases are used very differently. 
In pair 6, both the candidate and the reference contain \nlstring{the human dignity of the man}. 
Yet the two sentences convey very different meaning. 
\method agrees with the human judgment and ranks the pair low. 
In contrast,  \sentbleu considers the pair as relatively similar because of the significant word overlap. 

The bottom half of Table~\ref{tab:sup-example-table} shows examples where \method and human judgments disagree about the ranking. 
We observe that \method finds it difficult to detect factual errors. 
For example, \method assigns high similarity to pair 11 when the translation replaces \nlstring{German language} with \nlstring{English} and pair 12 where the translation incorrectly outputs \nlstring{Tuesday} when it is supposed to generate \nlstring{Monday}.
\method also fails to identify that \nlstring{5ft 11in} is equivalent with \nlstring{1.80 metres} in pair 18.
As a result, \method assigns low similarity to the eighth pair in Table~\ref{tab:sup-example-table}.
\sentbleu also suffers from these limitations.

\begin{figure}[t]
\centering
\includegraphics[width=\linewidth]{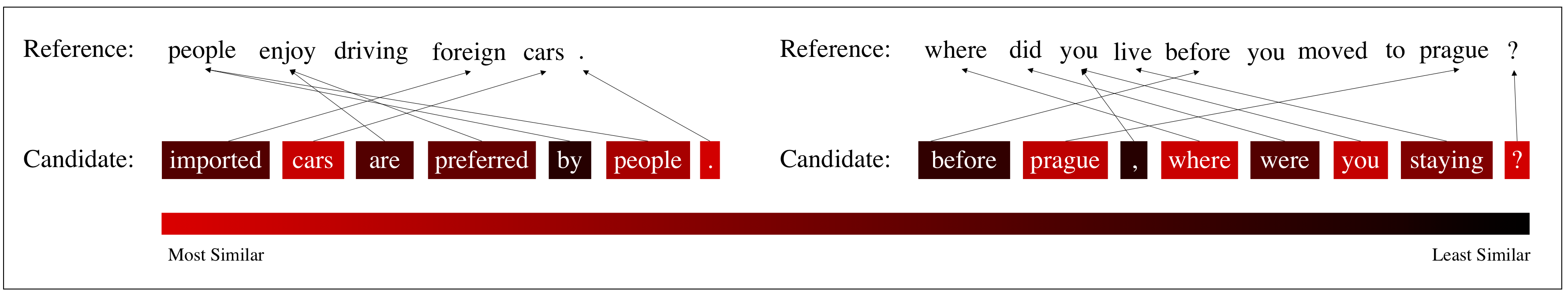}
\caption{\method visualization. The cosine similarity of each word matching in $\methodp{}$ are color-coded.}
\label{fig:color_example}
\end{figure}

Figure~\ref{fig:color_example} visualizes the \method matching of two pairs of candidate and reference sentences. 
The figure illustrates how $\methodf{}$ matches synonymous phrases, such as \nlstring{imported cars} and \nlstring{foreign cars}. 
We also see that $\methodf{}$ effectively matches words even given a high ordering distortion, for example the token \nlstring{people} in the figure.

\newpage

\section{Representation Choice}
\label{sec:bert-study}

\begin{figure}[h!]
\centering
\includegraphics[width=.48\textwidth]{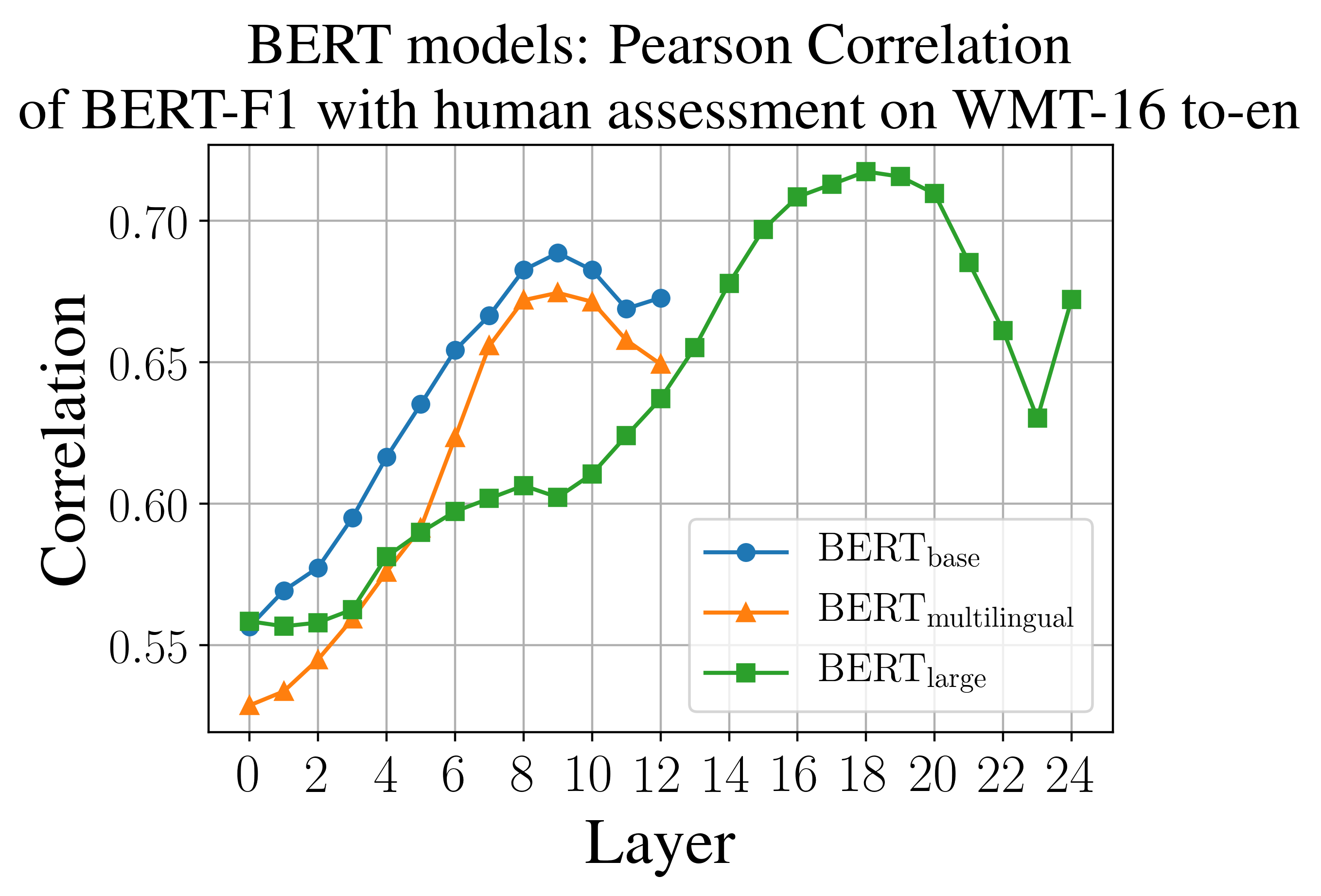}\quad
\includegraphics[width=.48\textwidth]{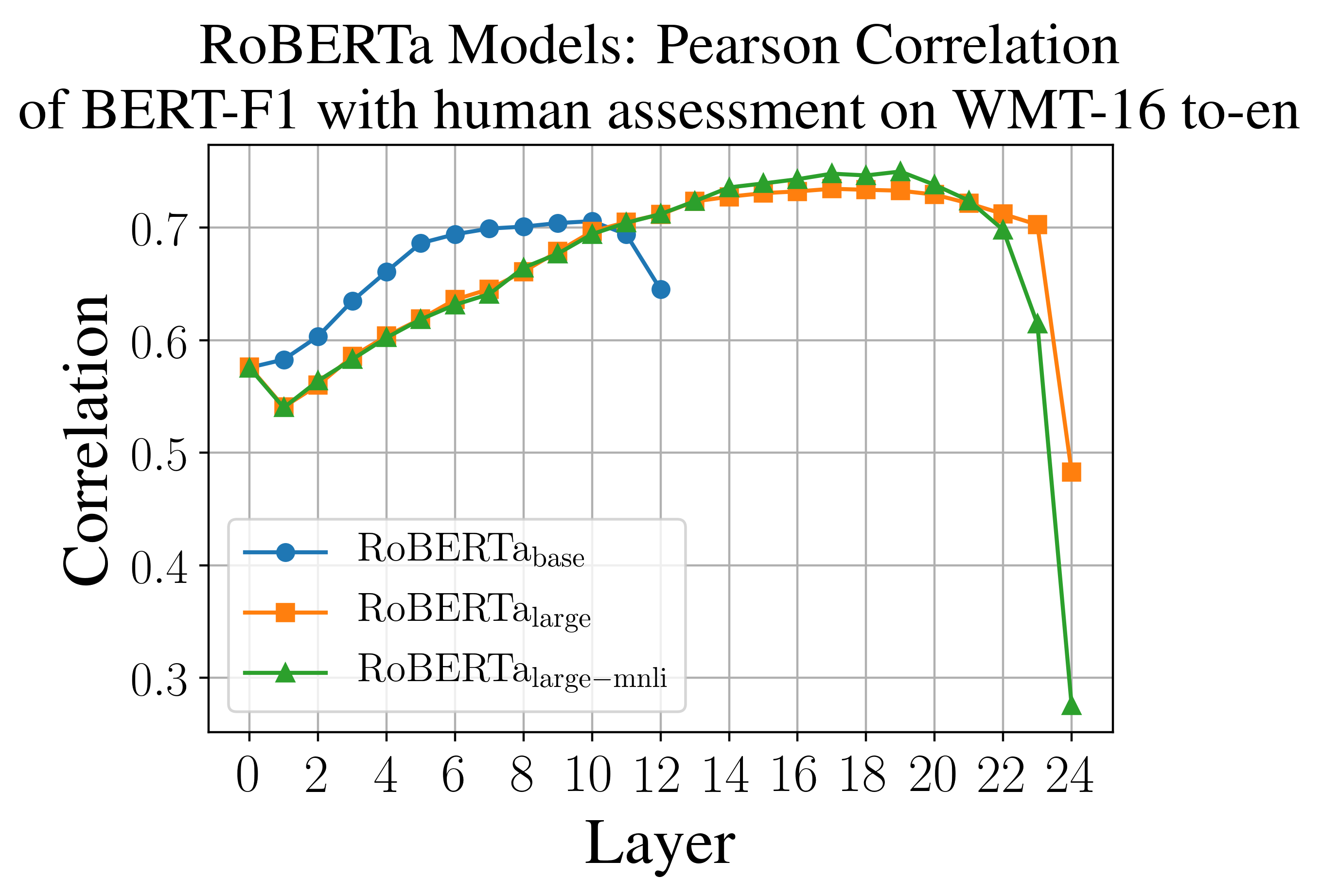}
\medskip
\includegraphics[width=.48\textwidth]{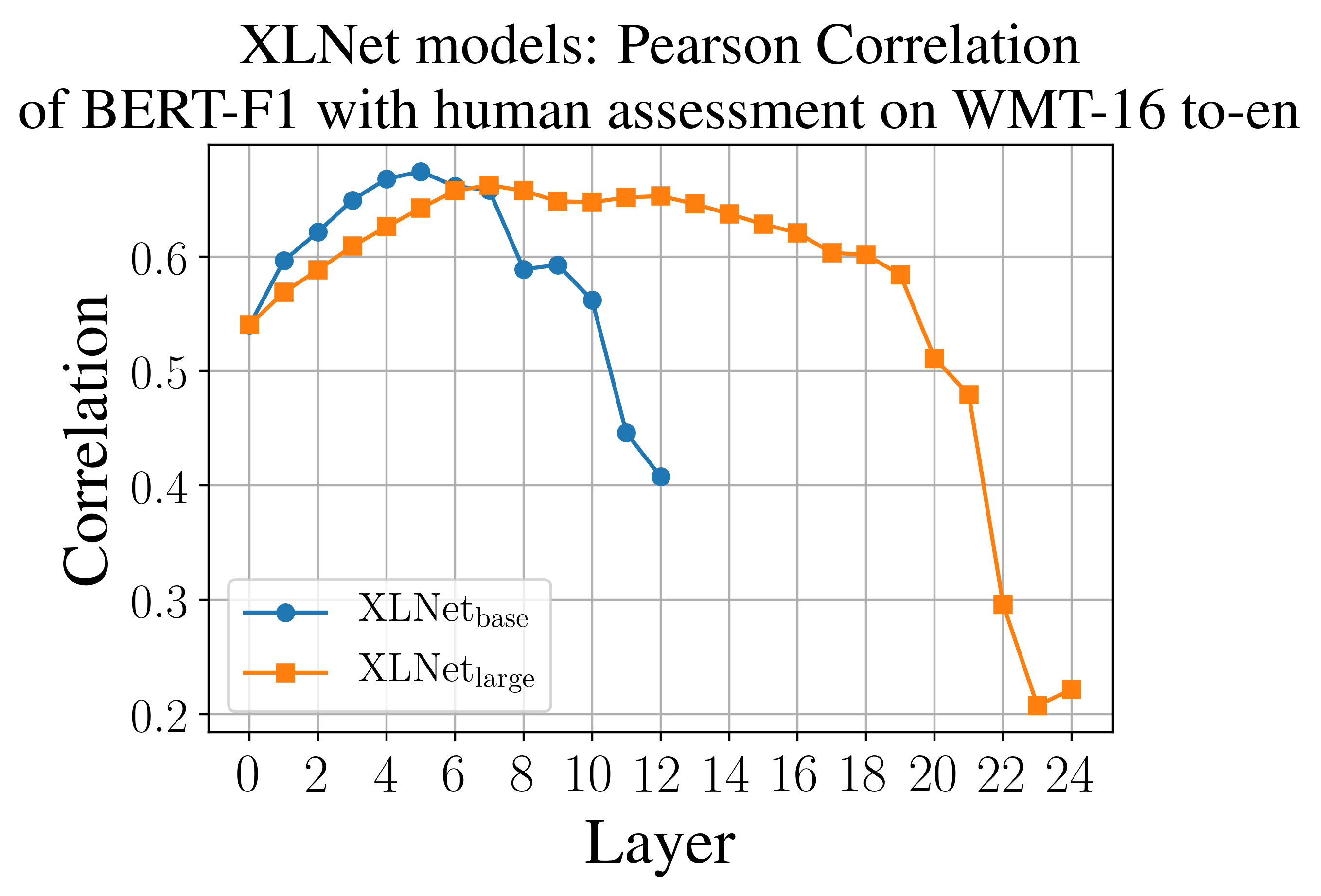}
\includegraphics[width=.48\textwidth]{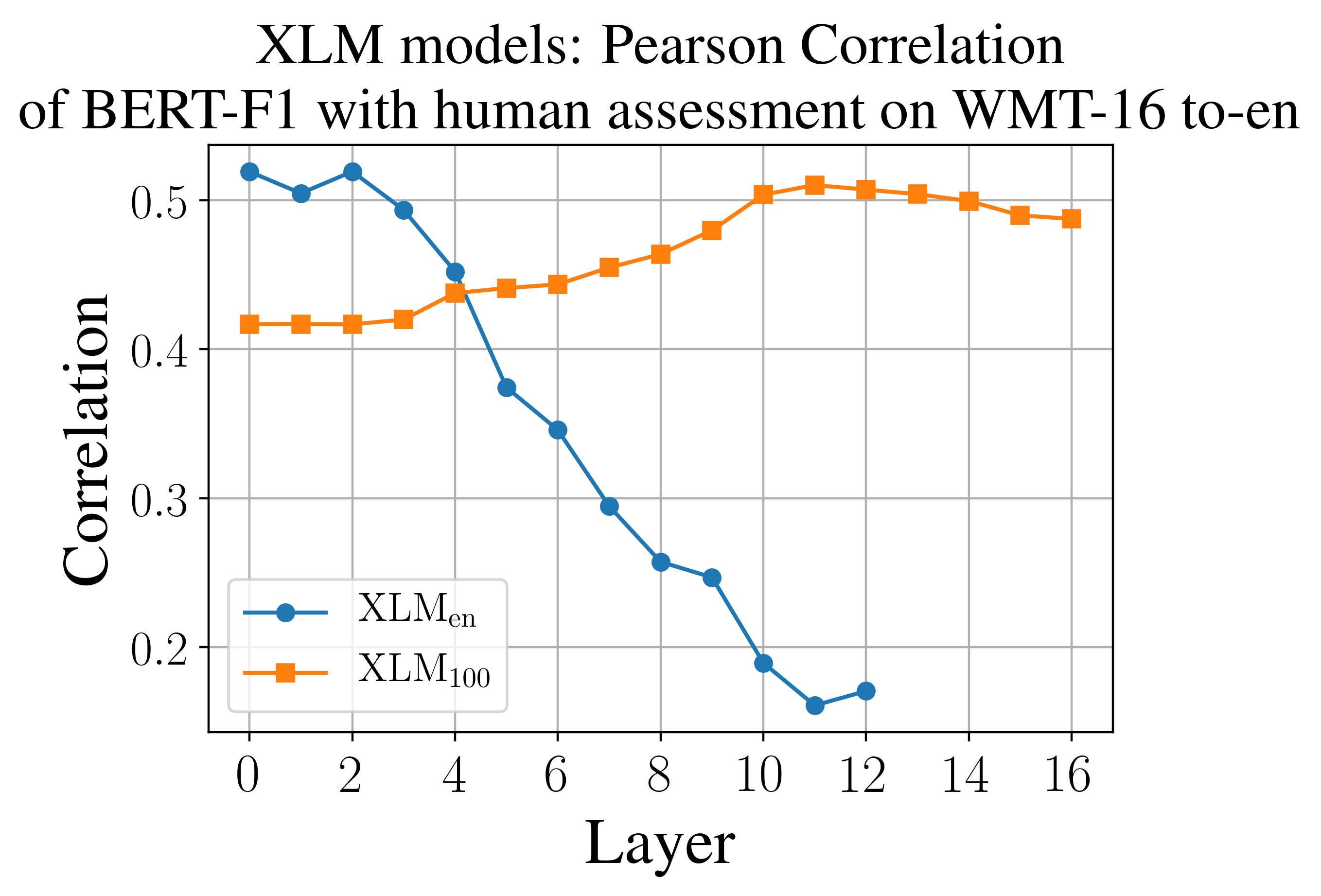}\quad
\medskip
\includegraphics[width=.48\textwidth]{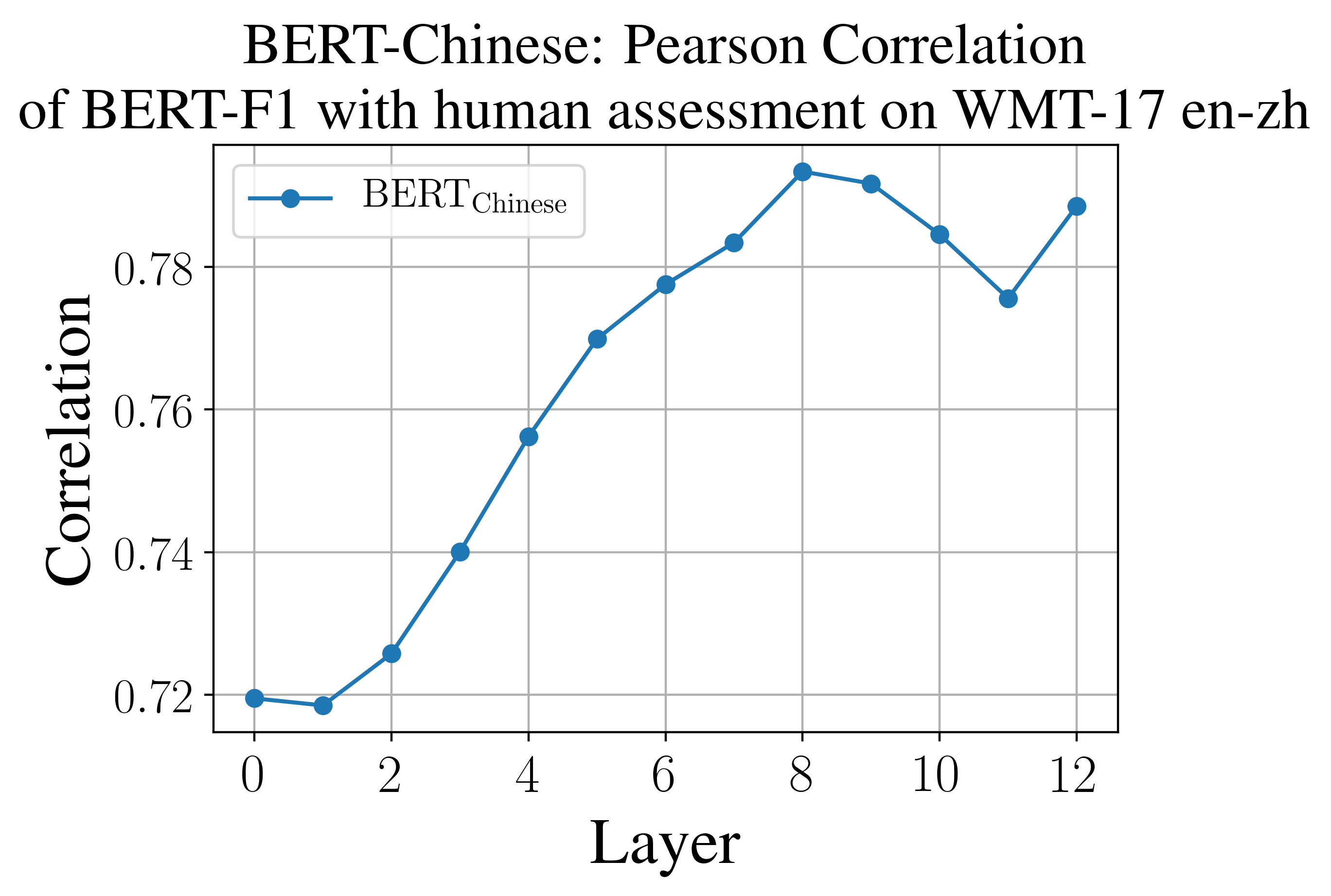}
\caption{Pearson correlation of $\methodf{}$ computed with different models, across different layers, with segment-level human judgments on the WMT16 to-English machine translation task. The WMT17 English-Chinese data is used for the BERT Chinese model. Layer 0 corresponds to using BPE embeddings. Consistently, correlation drops significantly in the final layers.}
\label{fig:layer}
\end{figure}

As suggested by previous works~\citep{elmo, alternate}, selecting a good layer or a good combination of layers from the BERT model is important.
In designing \method{}, we use WMT16 segment-level human judgment data as a development set to facilitate our representation choice. 
For Chinese models, we tune with the WMT17 ``en-zh'' data because the language pair ``en-zh'' is not available in WMT16.
In Figure~\ref{fig:layer}, we plot the change of human correlation of $\methodf$ over different layers of BERT, RoBERTa, XLNet and XLM models. 
Based on results from different models, we identify a common trend that $\methodf{}$ computed with the intermediate representations tends to work better. 
We tune the number of layer to use for a range of publicly available models.\footnote{\href{https://huggingface.co/pytorch-transformers/pretrained_models.html}{https://huggingface.co/pytorch-transformers/pretrained\_models.html}}
Table \ref{tab:best-layer} shows the results of our hyperparameter search.

\begin{table*}[h!]
    \footnotesize
    \centering
    
\begin{tabular}{c|c|c}
\toprule
Model & Total Number of Layers & Best Layer \\
\midrule
bert-base-uncased & 12 & 9 \\
bert-large-uncased & 24 & 18 \\
bert-base-cased-finetuned-mrpc & 12 & 9 \\
bert-base-multilingual-cased & 12 & 9 \\
bert-base-chinese & 12 & 8 \\
roberta-base & 12 & 10 \\
roberta-large & 24 & 17 \\
roberta-large-mnli & 24 & 19 \\
xlnet-base-cased & 12 & 5 \\
xlnet-large-cased & 24 & 7\\
xlm-mlm-en-2048 & 12 & 7 \\
xlm-mlm-100-1280 & 16 & 11\\
\bottomrule
\end{tabular}
    \vspace{-2pt}
    \caption{Recommended layer of representation to use for \method. The layers are chosen based on a held-out validation set (WMT16).}
    \label{tab:best-layer}
\end{table*}

\clearpage

\section{Ablation Study of \moverscore}
\label{sec:sup-mover-ablate}

Word Mover's Distance~\citep[WMD;][]{kusner2015word} is a semantic similarity metric that relies on word embeddings and optimal transport.
\moverscore~\citep{moverscore} combines contextual embeddings and WMD for text generation evaluation.
In contrast, \method adopts a greedy approach to aggregate token-level information.
In addition to using WMD for generation evaluation, \citet{moverscore} also introduce various other improvements. 
We do a detailed ablation study to understand the benefit of each improvement, and to investigate whether it can be applied to \method. 
We use a 12-layer uncased BERT model on the WMT17 to-English segment-level data, the same setting as \citet{moverscore}. 

We identify several differences between \moverscore and \method by analyzing the released source code. We isolate each difference, and mark it with a bracketed tag for our ablation study:
\begin{enumerate}
    \item  \texttt{[MNLI]} Use a BERT model fine-tuned on MNLI~\citep{mnli}.
    \item \texttt{[PMEANS]} Apply power means~\citep{pmeans} to aggregate the information of different layers.\footnote{ \citet{moverscore} uses the embeddings from the last five layers from BERT and L2-normalizes the embedding vectors at each layer before computing the P-MEANs and L2-normalizing the concatenated P-MEANS.} 
    \item \texttt{[IDF-L]} For reference sentences, instead of computing the $\idf$ scores on the 560 sentences in the segment-level data (\texttt{[IDF-S]}), compute the $\idf$ scores on the 3,005 sentences in the system-level data. 
    \item \texttt{[SEP]} For candidate sentences, recompute the $\idf$ scores on the candidate sentences. The weighting of reference tokens are kept the same as in \texttt{[IDF-S]}
    \item \texttt{[RM]} Exclude punctuation marks and sub-word tokens except the first sub-word in each word from the matching. 
\end{enumerate}

We follow the setup of \citet{moverscore} and use their released fine-tuned BERT model to conduct the experiments.
Table~\ref{tab:mover_ablate} shows the results of our ablation study. 
We report correlations for the two  variants of WMD \citet{moverscore} study:  unigrams (WMD1) and bigrams (WMD2).
Our $\methodf{}$ corresponds to the vanilla setting and the importance weighted variant corresponds to the \texttt{[IDF-S]} setting. 
The complete \moverscore metric corresponds to \texttt{[IDF-S]}+\texttt{[SEP]}+\texttt{[PMEANS]}+\texttt{[MNLI]}+\texttt{[RM]}.
We make several observations.
First, for all language pairs except fi-en and lv-en, we can replicate the reported performance. 
For these two language pairs,  \citet{moverscore} did not release their implementations at the time of publication.\footnote{A public comment on the project page indicates that some of the techniques are not applied for these two language pairs  (\href{https://github.com/AIPHES/emnlp19-moverscore/issues/1}{https://github.com/AIPHES/emnlp19-moverscore/issues/1}).}
Second, we confirm the effectiveness of \texttt{[PMEANS]} and \texttt{[MNLI]}. 
In Appendix~\ref{sec:sup-additiona-data}, we study more pre-trained models and further corroborate this conclusion.
However, the contribution of other techniques, including \texttt{[RM]} and \texttt{[SEP]}, seems less stable.
Third, replacing greedy matching with WMD does not lead to consistent improvement.
In fact, oftentimes \method is the better metric when given the same setup.
In general, for any given language pair, \method is always among the best performing ones. 
Given the current results, it is not clear tht WMD is  better than greedy matching for text generation evaluation.

\begin{table}[h!]
    \centering
    \footnotesize
    \setlength\tabcolsep{5pt}%
\begin{tabular}{c|cccccccc}
\toprule
Ablation & Metric & cs-en & de-en & fi-en & lv-en & ru-en & tr-en & zh-en\\
\midrule
\multirow{3}{*}{ \shortstack[c]{Vanilla} }
 & WMD1 & 0.628 & 0.655 & 0.795 & 0.692 & 0.701 & 0.715 & 0.699\\
 & WMD2 & 0.638 & 0.661 & 0.797 & 0.695 & 0.700 & 0.728 & 0.714\\
 & $F_{{\text{{BERT}}}}$ & 0.659 & 0.680 & 0.817 & 0.702 & 0.719 & 0.727 & 0.717\\
\midrule
\multirow{3}{*}{ \shortstack[c]{IDF-S} }
 & WMD1 & 0.636 & 0.662 & 0.824 & 0.709 & 0.716 & 0.728 & 0.713\\
 & WMD2 & 0.643 & 0.662 & 0.821 & 0.708 & 0.712 & 0.732 & 0.715\\
 & $F_{{\text{{BERT}}}}$ & 0.657 & 0.681 & 0.823 & 0.713 & 0.725 & 0.718 & 0.711\\
\midrule
\multirow{3}{*}{ \shortstack[c]{IDF-L} }
 & WMD1 & 0.633 & 0.659 & 0.825 & 0.708 & 0.716 & 0.727 & 0.715\\
 & WMD2 & 0.641 & 0.661 & 0.822 & 0.708 & 0.713 & 0.730 & 0.716\\
 & $F_{{\text{{BERT}}}}$ & 0.655 & 0.682 & 0.823 & 0.713 & 0.726 & 0.718 & 0.712\\
\midrule
\multirow{3}{*}{ \shortstack[c]{IDF-L + SEP} }
 & WMD1 & 0.651 & 0.660 & 0.819 & 0.703 & 0.714 & 0.724 & 0.715\\
 & WMD2 & 0.659 & 0.662 & 0.816 & 0.702 & 0.712 & 0.729 & 0.715\\
 & $F_{{\text{{BERT}}}}$ & 0.664 & 0.681 & 0.818 & 0.709 & 0.724 & 0.716 & 0.710\\
\midrule
\multirow{3}{*}{ \shortstack[c]{IDF-L + SEP\\ + RM} }
 & WMD1 & 0.651 & 0.686 & 0.803 & 0.681 & \textbf{0.730} & 0.730 & 0.720\\
 & WMD2 & 0.664 & 0.687 & 0.797 & 0.679 & \textbf{0.728} & 0.735 & 0.718\\
 & $F_{{\text{{BERT}}}}$ & 0.659 & 0.695 & 0.800 & 0.683 & \textbf{0.734} & 0.722 & 0.712\\
\midrule
\multirow{3}{*}{ \shortstack[c]{IDF-L + SEP\\ + PMEANS} }
 & WMD1 & 0.658 & 0.663 & 0.820 & 0.707 & 0.717 & 0.725 & 0.712\\
 & WMD2 & 0.667 & 0.665 & 0.817 & 0.707 & 0.717 & 0.727 & 0.712\\
 & $F_{{\text{{BERT}}}}$ & \textbf{0.671} & 0.682 & 0.819 & 0.708 & 0.725 & 0.715 & 0.704\\
\midrule
\multirow{3}{*}{ \shortstack[c]{IDF-L + SEP\\ + MNLI} }
 & WMD1 & 0.659 & 0.679 & 0.822 & 0.732 & 0.718 & 0.746 & 0.725\\
 & WMD2 & 0.664 & 0.682 & 0.819 & 0.731 & 0.715 & 0.748 & 0.722\\
 & $F_{{\text{{BERT}}}}$ & 0.668 & 0.701 & 0.825 & \textbf{0.737} & 0.727 & 0.744 & 0.725\\
\midrule
\multirow{3}{*}{ \shortstack[c]{IDF-L + SEP\\ + PMEANS + MNLI} }
 & WMD1 & 0.672 & 0.686 & \textbf{0.831} & \textbf{0.738} & 0.725 & 0.753 & \textbf{0.737}\\
 & WMD2 & \textbf{0.677} & 0.690 & 0.828 & \textbf{0.736} & 0.722 & \textbf{0.755} & 0.735\\
 & $F_{{\text{{BERT}}}}$ & \textbf{0.682} & 0.707 & \textbf{0.836} & \textbf{0.741} & 0.732 & 0.751 & \textbf{0.736}\\
\midrule
\multirow{3}{*}{ \shortstack[c]{IDF-L + SEP\\ + PMEANS + MNLI\\ + RM} }
 & WMD1 & 0.670 & 0.708 & 0.821 & 0.717 & \textbf{0.738} & \textbf{0.762} & \textbf{0.744}\\
 & WMD2 & \textbf{0.679} & \textbf{0.709} & 0.814 & 0.716 & \textbf{0.736} & \textbf{0.762} & 0.738\\
 & $F_{{\text{{BERT}}}}$ & \textbf{0.676} & \textbf{0.717} & 0.824 & 0.719 & \textbf{0.740} & \textbf{0.757} & \textbf{0.738}\\
\bottomrule
\end{tabular}
    \caption{Ablation Study of \moverscore and \method using Pearson correlations on the WMT17 to-English segment-level data. Correlations that are not outperformed by others for that language pair under Williams Test are bolded. We observe that using WMD does not consistently improve \method.}
    \label{tab:mover_ablate}
    \vspace{-10pt}
\end{table}

\clearpage

\section{Additional Experiments on Abstractive Text Compression}
\label{sec:sup-abs-text-comp}
We use the human judgments provided from the MSR Abstractive Text Compression Dataset~\citep{toutanova2016dataset} to illustrate the applicability of \method to abstractive text compression evaluation.
The data includes three types of human scores: (a) meaning: how well a compressed text preserve the meaning of the original text; (b) grammar: how grammatically correct a compressed text is; and (c) combined: the average of the meaning and the grammar scores.
We follow the experimental setup of \citet{toutanova2016dataset} and report Pearson correlation between \method and the three types of human scores.
Table~\ref{tab:msr-atcd} shows that $R_{\text{BERT}}$ has the highest correlation with human meaning judgments, and $P_{\text{BERT}}$ correlates highly with human grammar judgments. $F_{\text{BERT}}$ provides a balance between the two aspects.

\begin{table*}[t!]
    \centering
    \footnotesize
    \begin{tabular}{c|cccc}
\toprule
Type & Metric & Meaning & Grammar & Combined \\
\midrule 
\multirow{3}{*}{ \method }
& $P_{\text{BERT}}$ & 0.36 & 0.47 & 0.46 \\
& $R_{\text{BERT}}$ & 0.64 & 0.29 & 0.52 \\
& $F_{\text{BERT}}$ & 0.58 & 0.41 & 0.56 \\
\midrule
\multirow{4}{*}{ Common metrics }
& \bleu & 0.46 & 0.13 & 0.33 \\
& \metric{Meteor} & 0.53 & 0.11 & 0.36 \\
& \metric{ROUGE-L} & 0.51 & 0.16 & 0.38 \\
& \metric{SARI} & 0.50 & 0.15 & 0.37 \\
\midrule
\multirow{3}{*}{ \shortstack[c]{Best metrics according to \\ \citet{toutanova2016dataset} }}
& \metric{SKIP-2+Recall+MULT-PROB} & 0.59 & N/A & 0.51 \\
& \metric{PARSE-2+Recall+MULT-MAX} & N/A & 0.35 & 0.52 \\
& \metric{PARSE-2+Recall+MULT-PROB} & 0.57 & 0.35 & 0.52 \\
\bottomrule
\end{tabular}
    \caption{Pearson correlations with human judgments on the MSR Abstractive Text Compression Dataset.}
    \label{tab:msr-atcd}
\end{table*}

\clearpage

\section{\protect\method{} of Recent MT Models}
\label{sec:sup-recent-mt-score}

\begin{table*}[t!]
    \footnotesize
    % \gianttablefont
    \centering
    \setlength{\tabcolsep}{3pt}
    \begin{tabular}{c|l|l|lll|lll}
    \toprule
    Task & Model & \bleu & $\hatmethodp{}$ & $\hatmethodr{}$ & $\hatmethodf{}$ & $\methodp{}$ & $\methodr{}$ & $\methodf{}$ \\
    \midrule
    \multirow{3}{*}{ \shortstack[l]{WMT14 \\ En-De} } & ConvS2S \citep{gehring2017convs2s} & 0.266 & 0.6099 & 0.6055 & 0.6075 & 0.8499 & 0.8482 & 0.8488 \\
    & Transformer-big$^{**}$ \citep{snmt} &  \textbf{0.298}  & \textbf{0.6587} & \textbf{0.6528} & \textbf{0.6558} & \textbf{0.8687} & \textbf{0.8664} & \textbf{0.8674} \\
    & DynamicConv$^{***}$ \citep{wu2018pay} & 0.297 & 0.6526 & 0.6464 & 0.6495  &0.8664 & 0.8640 & 0.8650 \\
    \midrule
    \multirow{3}{*}{ \shortstack[l]{WMT14 \\ En-Fr} } & ConvS2S \citep{gehring2017convs2s} & 0.408 & 0.6998 & 0.6821 & 0.6908 & 0.8876 & 0.8810 & 0.8841 \\
    & Transformer-big \citep{snmt} & \textbf{0.432} & 0.7148 & 0.6978 & 0.7061 & 0.8932 & 0.8869 & 0.8899 \\
    & DynamicConv \citep{wu2018pay} & \textbf{0.432} & \textbf{0.7156} & \textbf{0.6989} & \textbf{0.7071} & \textbf{0.8936} & \textbf{0.8873} &\textbf{0.8902} \\
    \midrule
    \multirow{3}{*}{ \shortstack[l]{IWSLT14 \\ De-En} }
    & Transformer-iwslt$^+$ \citep{ott2019fairseq} & 0.350 & 0.6749 & 0.6590 & 0.6672 & 0.9452 & 0.9425 & 0.9438 \\
    & LightConv ~\citep{wu2018pay} & 0.348 & 0.6737 & 0.6542 & 0.6642 & 0.9450 & 0.9417 & 0.9433 \\
    & DynamicConv~\citep{wu2018pay}  & \textbf{0.352} & \textbf{0.6770} & \textbf{0.6586} & \textbf{0.6681} & \textbf{0.9456} & \textbf{0.9425} & \textbf{0.9440} \\
    \bottomrule
    \end{tabular}
    
    \caption{
    \bleu scores and \method{}s of publicly available pre-trained MT models in 
    fairseq~\citep{ott2019fairseq}.
    We show both rescaled scores marked with $\hat{\protect\phantom{p}}$ and raw \method{}s. $^*$: trained on unconfirmed WMT data version, $^{**}$: trained on WMT16 + ParaCrawl, $^{***}$: trained on WMT16, $^+$: trained by us using fairseq.}
    \label{tab:mt_performance}
\end{table*}

\autoref{tab:mt_performance} shows the \bleu scores and the \method{}s of pre-trained machine translation models on WMT14 English-to-German, WMT14 English-to-French, IWSLT14 German-to-English task. We used publicly available pre-trained models from 
fairseq~\citep{ott2019fairseq}.\footnote{
Code and pre-trained model available at \href{https://github.com/pytorch/fairseq}{https://github.com/pytorch/fairseq}.}
Because a pre-trained Transformer model on IWSLT is not released, we trained our own using the fairseq library.
We use multilingual cased $\bertbase$\footnote{Hash code: \texttt{\scriptsize bert-base-multilingual-cased\_L9\_version=0.2.0}} for English-to-German and English-to-French pairs, and English uncased $\bertbase$\footnote{Hash code: \texttt{\scriptsize roberta-large\_L17\_version=0.2.0}} for German-to-English pairs.
Interestingly, the gap between a DynamicConv~\citep{wu2018pay} trained on only WMT16 and a Transformer~\citep{snmt} trained on WMT16 and ParaCrawl\footnote{\href{http://paracrawl.eu/download.html}{http://paracrawl.eu/download.html}} (about 30$\times$ more training data) becomes larger when evaluated with \method rather than \bleu.

\clearpage

\section{Additional Results}
\label{sec:sup-additiona-data}

In this section, we present additional experimental results:
\begin{enumerate}
\item Segment-level and system-level correlation studies on three years of WMT metric evaluation task (WMT16--18) 
\item Model selection study on WMT18 10K hybrid systems
\item System-level correlation study on 2015 COCO captioning challenge
\item Robustness study on PAWS-QQP.
\end{enumerate}

Following BERT~\citep{bert}, a variety of Transformer-based~\citep{transformer} pre-trained contextual embeddings have been proposed and released.
We conduct additional experiments with four types of pre-trained embeddings: BERT, XLM~\citep{xlm}, XLNet~\citep{yang2019xlnet}, and RoBERTa~\citep{roberta}.
XLM (Cross-lingual Language Model) is a Transformer pre-trained on the translation language modeling of predicting masked tokens from a pair of sentence in two different languages and masked language modeling tasks using multi-lingual training data.
\citet{yang2019xlnet} modify the Transformer architecture and pre-train it on a permutation language modeling task resulting in some improvement on top of the original BERT when fine-tuned on several downstream tasks.
\citet{roberta} introduce RoBERTa (Robustly optimized BERT approach) and demonstrate that an optimized BERT model is comparable to or sometimes outperforms an XLNet on downstream tasks.

We perform a comprehensive study with the following pre-trained contextual embedding models:\footnote{Denoted by names specified at  \href{https://huggingface.co/pytorch-transformers/pretrained_models.html}{https://huggingface.co/pytorch-transformers/pretrained\_models.html}.}
\begin{itemize}
    \item BERT models: \texttt{bert-base-uncased}, \texttt{bert-large-uncased}, \texttt{bert-based-chinese}, \texttt{bert-base-multilingual-cased}, and \texttt{bert-base-cased-mrpc}
    \item RoBERTa models:  \texttt{roberta-base},
    \texttt{roberta-large}, and \texttt{roberta-large-mnli}
    \item XLNet models: \texttt{xlnet-base-cased} and
    \texttt{xlnet-base-large}
    \item XLM models: \texttt{xlm-mlm-en-2048} and \texttt{xlm-mlm-100-1280}
\end{itemize}

\subsection{WMT Correlation Study}
\paragraph{Experimental setup}
Because of missing data in the released WMT16 dataset~\citep{wmt16em}, we are only able to experiment with to-English segment-level data, which contains the outputs of 50 different systems on 6 language pairs.
We use this data as the validation set for hyperparameter tuning (Appendix~\ref{sec:bert-study}).
Table~\ref{tab:sup-wmt16-to-seg} shows the Pearson correlations of all participating metrics and \method{}s computed with different pre-trained models. 
Significance testing for this dataset does not include  the baseline metrics because the released dataset does not contain the original outputs from the baseline metrics. We  conduct significance testing between \method results only.

The WMT17 dataset~\citep{wmt17em} contains outputs of 152 different translations on 14 language pairs. 
We experiment on the segment-level and system-level data on both to-English and from-English language pairs.
We exclude fi-en data from the segment-level experiment due to an error in the released data.
We compare our results to all participating metrics and perform standard significance testing as done by \cite{wmt17em}. 
Tables~\ref{tab:sup-wmt17-to-seg}--\ref{tab:sup-wmt17-from-sys} show the results. 

The WMT18 dataset~\citep{wmt18em} contains outputs of 159 translation systems on 14 language pairs.
In addition to the results in Tables~\ref{tab:wmt18-sys}--\ref{tab:wmt18-seg}, we complement the study with the correlations of all participating metrics in WMT18 and results from using different contextual models for \method.

\paragraph{Results}
Table~\ref{tab:sup-wmt16-to-seg}--\ref{tab:sup-wmt18-from-sys-hybrids} collectively showcase the effectiveness of \method in correlating with human judgments.
The improvement of \method is more pronounced on the segment-level than on the system-level.
We also see that more optimized or larger BERT models can produce better contextual representations (e.g., comparing $F_{\text{RoBERTa--Large}}$ and $F_{\text{BERT--Large}}$).
In contrast, the smaller XLNet performs better than a large one.
Based on the evidence  in Figure~\ref{tab:best-layer} and Tables~\ref{tab:sup-wmt16-to-seg}--\ref{tab:sup-wmt18-from-sys-hybrids}, we hypothesize that the permutation language task, though leading to a good set of model weights for fine-tuning on downstream tasks, does not necessarily produce informative pre-trained embeddings for generation evaluation. 
We also observe that fine-tuning pre-trained models on a related task, such as natural language inference~\citep{mnli}, can lead to better human correlation in evaluating text generation.
Therefore, for evaluating English sentences, we recommend computing \method with a 24-layer RoBERTa model fine-tuned on the MNLI dataset.
For evaluating Non-English sentences, both the multilingual BERT model and the XLM model trained on 100 languages are suitable candidates.
We also recommend using domain- or language-specific contextual embeddings when possible, such as using BERT Chinese models for evaluating Chinese tasks.
In general, we advise users to consider the target domain and languages when selecting the exact configuration to use.

\subsection{Model Selection Study}

\paragraph{Experimental setup} 

Similar to Section~\ref{sec:exp}, we use the 10K hybrid systems super-sampled from WMT18. 
We randomly select 100 out of 10K hybrid systems, rank them using automatic metrics, and repeat this process 100K times. 
We add to the results in the main paper (Table~\ref{tab:wmt18-model-select}) performance of all participating metrics in WMT18 and results from using different contextual embedding models for \method.
We reuse the hybrid configuration and metric outputs released in WMT18.
In addition to the Hits@1 measure, we evaluate the metrics using (a) mean reciprocal rank (MRR) of the top metric-rated system in human rankings, and (b) the absolute human score difference (Diff) between the top metric- and human-rated systems.
Hits@1 captures a metric's ability to select the best system.
The other two measures quantify the amount of error a metric makes in the selection process. 
Tables~\ref{tab:sup-wmt18-to-modelselect-hit1}--\ref{tab:sup-wmt18-from-modelselect-diff} show the results from these experiments. 

\paragraph{Results} 
The additional results further support our conclusion from Table~\ref{tab:wmt18-model-select}: \method demonstrates better model selection performance.
We also observe that the supervised metric RUSE displays strong model selection ability.

\subsection{Image Captioning on COCO}
We follow the  experimental setup described in Section~\ref{sec:exp}.
Table~\ref{tab:coco_appendix} shows the correlations of several pre-trained contextual embeddings.
We observe that precision-based methods such as \bleu and $\methodp$ are weakly correlated with human judgments on image captioning tasks.
We hypothesize that this is because human judges prefer captions that capture the main objects in a picture for image captioning.
In general, $\methodr$ has a high correlation, even surpassing the task-specific metric \metric{Spice}~\cite{spice}.
While the fine-tuned RoBERTa-Large model does not result in the highest correlation, it is one of the best metrics.

\subsection{Robustness Analysis on PAWS-QQP}
We present the full results of the robustness study described in Section~\ref{sec:robust} in Table~\ref{tab:paws_appendix}. 
In general, we observe that \method is more robust than other commonly used metrics.
\method computed with the 24-layer RoBERTa model performs the best.
Fine-tuning RoBERTa-Large on MNLI~\citep{mnli} can significantly improve the robustness against adversarial sentences. However, a fine-tuned BERT on MRPC (Microsoft Research Paraphrasing Corpus)~\citep{mrpc} performs worse than its counterpart.

\newpage
\begin{table*}[t!]
    \centering
    \gianttablefont
\begin{tabular}{c|ccccccc}
\toprule
Setting & Metric & cs-en & de-en & fi-en & ro-en & ru-en & tr-en\\
 &  & 560 & 560 & 560 & 560 & 560 & 560\\
\midrule
\multirow{14}{*}{Unsupervised}
& \textsc{DPMFcomb} & 0.713 & 0.584 & 0.598 & 0.627 & 0.615 & 0.663 \\
& \textsc{metrics-f} & 0.696 & 0.601 & 0.557 & 0.662 & 0.618 & 0.649 \\
& \textsc{cobalt-f.} & 0.671 & 0.591 & 0.554 & 0.639 & 0.618 & 0.627 \\
& \textsc{upf-coba.} & 0.652 & 0.550 & 0.490 & 0.616 & 0.556 & 0.626 \\
& \textsc{MPEDA} & 0.644 & 0.538 & 0.513 & 0.587 & 0.545 & 0.616 \\
& \textsc{chrF2} & 0.658 & 0.457 & 0.469 & 0.581 & 0.534 & 0.556 \\
\bottomrule
\end{tabular}
    \caption{Pearson correlations with segment-level human judgments on WMT16 to-English translations. Correlations of metrics not significantly outperformed by any other for that language pair are highlighted in bold. For each language pair, we specify the number of examples.}
    \label{tab:sup-wmt16-to-seg}
\end{table*}

\newpage
\begin{table*}[t!]
    \centering
    \gianttablefont
    \begin{tabular}{c|cccccccc}
\toprule
Setting & Metric & cs-en & de-en & fi-en & lv-en & ru-en & tr-en & zh-en\\
 &  & 560 & 560 & 560 & 560 & 560 & 560 & 560\\
\midrule
\multirow{7}{*}{Unsupervised}
 & \textsc{chrF}\xspace & 0.514 & 0.531 & 0.671 & 0.525 & 0.599 & 0.607 & 0.591\\
 & \textsc{chrF++}\xspace & 0.523 & 0.534 & 0.678 & 0.520 & 0.588 & 0.614 & 0.593\\
 & \textsc{MEANT 2.0}\xspace & 0.578 & 0.565 & 0.687 & 0.586 & 0.607 & 0.596 & 0.639\\
 & \textsc{MEANT 2.0-nosrl}\xspace & 0.566 & 0.564 & 0.682 & 0.573 & 0.591 & 0.582 & 0.630\\
 & \textsc{SentBLEU}\xspace & 0.435 & 0.432 & 0.571 & 0.393 & 0.484 & 0.538 & 0.512\\
 & \textsc{TreeAggreg}\xspace & 0.486 & 0.526 & 0.638 & 0.446 & 0.555 & 0.571 & 0.535\\
\bottomrule
\end{tabular}
    \caption{Absolute Pearson correlations with segment-level human judgments on WMT17 to-English translations. Correlations of metrics not significantly outperformed by any other for that language pair are highlighted in bold. For each language pair, we specify the number of examples.}
    \label{tab:sup-wmt17-to-seg}
\end{table*}

\newpage
\begin{table*}[t!]
    \centering
    \footnotesize
    \begin{tabular}{c|cccccccc}
\toprule
Setting & Metric & en-cs & en-de & en-fi & en-lv & en-ru & en-tr & en-zh \\
&  & 32K & 3K & 3K & 3K & 560 & 247 & 560 \\
 &  & $\tau$ & $\tau$ & $\tau$ & $\tau$ & $|r|$ & $\tau$ & $|r|$ \\
\midrule
\multirow{9}{*}{Unsupervised}
& \textsc{AutoDA} & 0.041 & 0.099 & 0.204 & 0.130 & 0.511 & 0.409 & 0.609 \\
& \textsc{AutoDA-TECTO} & 0.336 & - & - & - & - & - & - \\
& \textsc{chrF} & 0.376 & 0.336 & 0.503 & 0.420 & 0.605 & 0.466 & 0.608 \\
& \textsc{chrF+} & 0.377 & 0.325 & 0.514 & 0.421 & 0.609 & 0.474 & - \\
& \textsc{chrF++} & 0.368 & 0.328 & 0.484 & 0.417 & 0.604 & 0.466 & 0.602 \\
\bottomrule
\end{tabular}
    \caption{Absolute Pearson correlation ($|r|$) and Kendall correlation ($\tau$) with segment-level human judgments on WMT17 from-English translations. Correlations of metrics not significantly outperformed by any other for that language pair are highlighted in bold. For each language pair, we specify the number of examples.}
    \label{tab:sup-wmt17-from-seg}
\end{table*}

\newpage
\begin{table*}[t!]
    \centering
    \gianttablefont
    \begin{tabular}{c|cccccccc}
\toprule
Setting & Metric & cs-en & de-en & fi-en & lv-en & ru-en & tr-en & zh-en\\
 &  & 4 & 11 & 6 & 9 & 9 & 10 & 16\\
\midrule
\multirow{13}{*}{Unsupervised}
 & \textsc{BLEU}\xspace & 0.971 & 0.923 & 0.903 & 0.979 & 0.912 & 0.976 & 0.864\\
 & \textsc{CDER}\xspace & 0.989 & 0.930 & 0.927 & 0.985 & 0.922 & 0.973 & 0.904\\
 & \textsc{CharacTER}\xspace & 0.972 & 0.974 & 0.946 & 0.932 & \textbf{0.958} & 0.949 & 0.799\\
 & \textsc{chrF}\xspace & 0.939 & 0.968 & 0.938 & 0.968 & 0.952 & 0.944 & 0.859\\
 & \textsc{chrF++}\xspace & 0.940 & 0.965 & 0.927 & 0.973 & 0.945 & 0.960 & 0.880\\
 & \textsc{MEANT 2.0}\xspace & 0.926 & 0.950 & 0.941 & 0.970 & \textbf{0.962} & 0.932 & 0.838\\
\bottomrule
\end{tabular}
    \caption{Absolute Pearson correlations with system-level human judgments on WMT17 to-English translations. Correlations of metrics not significantly outperformed by any other for that language pair are highlighted in bold. For each language pair, we specify the number of systems.}
    \label{tab:sup-wmt17-to-sys}
\end{table*}

\newpage
\begin{table*}[t!]
    \centering
    \footnotesize
    \begin{tabular}{c|ccccccc}
\toprule
Setting & Metric & en-cs & en-de & en-lv & en-ru & en-tr & en-zh\\
 &  & 14 & 16 & 17 & 9 & 8 & 11\\
\midrule
\multirow{13}{*}{Unsupervised}
 & \textsc{BLEU}\xspace & 0.956 & 0.804 & 0.866 & 0.898 & 0.924 & --\\
 & \textsc{CDER}\xspace & 0.968 & 0.813 & 0.930 & 0.924 & 0.957 & --\\
 & \textsc{CharacTER}\xspace & \textbf{0.981} & \textbf{0.938} & 0.897 & 0.939 & \textbf{0.975} & 0.933\\
 & \textsc{chrF}\xspace & \textbf{0.976} & 0.863 & \textbf{0.955} & 0.950 & \textbf{0.991} & 0.976\\
 & \textsc{chrF++}\xspace & \textbf{0.974} & 0.852 & \textbf{0.956} & 0.945 & 0.986 & 0.976\\
 & \textsc{MEANT 2.0}\xspace & -- & 0.858 & -- & -- & -- & 0.956\\
\bottomrule
\end{tabular}
    \caption{Absolute Pearson correlations with system-level human judgments on WMT17 from-English translations. Correlations of metrics not significantly outperformed by any other for that language pair are highlighted in bold. For each language pair, we specify the number of systems.}
    \label{tab:sup-wmt17-from-sys}
\end{table*}

\begin{table*}[t!]
    \centering
    \gianttablefont
    \begin{tabular}{c|cccccccc}
\toprule
Setting & Metric & cs-en & de-en & et-en & fi-en & ru-en & tr-en & zh-en \\
&  & 5K & 78K & 57K & 16K & 10K & 9K & 33K \\
\midrule
\multirow{8}{*}{Unsupervised}
 & \textsc{CharacTER} & 0.256 & 0.450 & 0.286 & 0.185 & 0.244 & 0.172 & 0.202 \\
 & \textsc{ITER} & 0.198 & 0.396 & 0.235 & 0.128 & 0.139 & -0.029 & 0.144 \\
 & \textsc{Meteor++} & 0.270 & 0.457 & 0.329 & 0.207 & 0.253 & 0.204 & 0.179 \\
 & \textsc{SentBLEU} & 0.233 & 0.415 & 0.285 & 0.154 & 0.228 & 0.145 & 0.178 \\
 & \textsc{UHH\_TSKM} & 0.274 & 0.436 & 0.300 & 0.168 & 0.235 & 0.154 & 0.151 \\
 & \textsc{YiSi-0} & 0.301 & 0.474 & 0.330 & 0.225 & 0.294 & 0.215 & 0.205 \\
\bottomrule
\end{tabular}
    \caption{Kendall correlations with segment-level human judgments on WMT18 to-English translations. Correlations of metrics not significantly outperformed by any other for that language pair are highlighted in bold. For each language pair, we specify the number of examples.}
    \label{tab:sup-wmt18-to-seg}
\end{table*}

\newpage
\begin{table*}[t!]
    \centering
    \footnotesize
    \begin{tabular}{c|cccccccc}
\toprule
Setting & Metric & en-cs & en-de & en-et & en-fi & en-ru & en-tr & en-zh\\
& & 5K & 20K & 32K & 10K & 22K & 1K & 29K \\
\midrule
\multirow{6}{*}{Unsupervised}
 & \textsc{CharacTER} & 0.414 & 0.604 & 0.464 & 0.403 & 0.352 & 0.404 & 0.313\\
 & \textsc{ITER} & 0.333 & 0.610 & 0.392 & 0.311 & 0.291 & 0.236 & - \\
 & \textsc{SentBLEU} & 0.389 & 0.620 & 0.414 & 0.355 & 0.330 & 0.261 & 0.311\\
 & \textsc{YiSi-0} & 0.471 & 0.661 & 0.531 & 0.464 & 0.394 & 0.376 & 0.318\\
 & \textsc{YiSi-1} & 0.496 & 0.691 & 0.546 & 0.504 & 0.407 & 0.418 & 0.323\\
 & \textsc{YiSi-1 srl} & - & 0.696 & - & - & - & - & 0.310 \\
\bottomrule
\end{tabular}
    \caption{Kendall correlations with segment-level human judgments on WMT18 from-English translations. Correlations of metrics not significantly outperformed by any other for that language pair are highlighted in bold. For each language pair, we specify the number of examples.}
    \label{tab:sup-wmt18-from-seg}
\end{table*}

\begin{table*}[t!]
    \centering
    \gianttablefont
    \begin{tabular}{c|cccccccc}
\toprule
Setting & Metric & cs-en & de-en & et-en & fi-en & ru-en & tr-en & zh-en\\
 & & 5 & 16 & 14 & 9 & 8 & 5 & 14\\
\midrule
\multirow{13}{*}{Unsupervised}
 & \textsc{BLEU}\xspace & 0.970 & 0.971 & \textbf{0.986} & 0.973 & 0.979 & 0.657 & 0.978\\
 & \textsc{CDER}\xspace & 0.972 & 0.980 & \textbf{0.990} & 0.984 & 0.980 & 0.664 & 0.982\\
 & \textsc{CharacTER}\xspace & 0.970 & 0.993 & 0.979 & 0.989 & 0.991 & 0.782 & 0.950\\
 & \textsc{ITER}\xspace & 0.975 & 0.990 & 0.975 & \textbf{0.996} & 0.937 & \textbf{0.861} & 0.980\\
 & \textsc{Meteor++}\xspace & 0.945 & 0.991 & 0.978 & 0.971 & \textbf{0.995} & 0.864 & 0.962\\
 & \textsc{NIST}\xspace & 0.954 & 0.984 & 0.983 & 0.975 & 0.973 & \textbf{0.970} & 0.968\\
\bottomrule
\end{tabular}
    \caption{Absolute Pearson correlations with system-level human judgments on WMT18 to-English translations. Correlations of metrics not significantly outperformed by any other for that language pair are highlighted in bold. For each language pair, we specify the number of systems.}
    \label{tab:sup-wmt18-to-sys}
\end{table*}

\newpage
\begin{table*}[t!]
    \centering
    \footnotesize
    \begin{tabular}{c|cccccccc}
\toprule
Setting & Metric & en-cs & en-de & en-et & en-fi & en-ru & en-tr & en-zh\\
 & & 5 & 16 & 14 & 12 & 9 & 8 & 14\\
\midrule
\multirow{13}{*}{Unsupervised}
 & \textsc{BLEU}\xspace & 0.995 & 0.981 & 0.975 & 0.962 & 0.983 & 0.826 & 0.947\\
 & \textsc{CDER}\xspace & 0.997 & 0.986 & 0.984 & 0.964 & 0.984 & 0.861 & 0.961\\
 & \textsc{CharacTER}\xspace & \textbf{0.993} & \textbf{0.989} & 0.956 & \textbf{0.974} & 0.983 & 0.833 & \textbf{0.983}\\
 & \textsc{ITER}\xspace & 0.915 & 0.984 & 0.981 & 0.973 & 0.975 & 0.865 & --\\
 & \textsc{Meteor++}\xspace & -- & -- & -- & -- & -- & -- & --\\
 & \textsc{NIST}\xspace & \textbf{0.999} & 0.986 & \textbf{0.983} & 0.949 & \textbf{0.990} & 0.902 & 0.950\\
\bottomrule
\end{tabular}
    \caption{Absolute Pearson correlations with system-level human judgments on WMT18 from-English translations. Correlations of metrics not significantly outperformed by any other for that language pair are highlighted in bold. For each language pair, we specify the number of systems.}
    \label{tab:sup-wmt18-from-sys}
\end{table*}

\newpage
\begin{table*}[t!]
    \centering
    \gianttablefont
    \begin{tabular}{c|cccccccc}
\toprule
Setting & Metric & cs-en & de-en & et-en & fi-en & ru-en & tr-en & zh-en\\
 &  & 10K & 10K & 10K & 10K & 10K & 10K & 10K\\
\midrule
\multirow{13}{*}{Unsupervised}
 & \textsc{BLEU}\xspace & 0.956 & 0.969 & 0.981 & 0.962 & 0.972 & 0.586 & 0.968\\
 & \textsc{CDER}\xspace & 0.964 & 0.980 & 0.988 & 0.976 & 0.974 & 0.577 & 0.973\\
 & \textsc{CharacTER}\xspace & 0.960 & 0.992 & 0.975 & 0.979 & 0.984 & 0.680 & 0.942\\
 & \textsc{ITER}\xspace & 0.966 & 0.990 & 0.975 & 0.989 & 0.943 & 0.742 & 0.978\\
 & \textsc{Meteor++}\xspace & 0.937 & 0.990 & 0.975 & 0.962 & 0.989 & 0.787 & 0.954\\
 & \textsc{NIST}\xspace & 0.942 & \textbf{0.982} & 0.980 & 0.965 & 0.965 & 0.862 & 0.959\\
\bottomrule
\end{tabular}
    \caption{Absolute Pearson correlations with human judgments on WMT18 to-English language pairs for 10K hybrid systems. Correlations of metrics not significantly outperformed by any other for that language pair are highlighted in bold. For each language pair, we specify the number of systems.}
    \label{tab:sup-wmt18-to-sys-hybrids}
\end{table*}

\newpage
\begin{table*}[t!]
    \centering
    \footnotesize
    \begin{tabular}{c|cccccccc}
\toprule
Setting & Metric & en-cs & en-de & en-et & en-fi & en-ru & en-tr & en-zh\\
 &  & 10K & 10K & 10K & 10K & 10K & 10K & 10K\\
\midrule
\multirow{13}{*}{Unsupervised}
 & \textsc{BLEU}\xspace & 0.993 & 0.977 & 0.971 & 0.958 & 0.977 & 0.796 & 0.941\\
 & \textsc{CDER}\xspace & 0.995 & 0.984 & 0.981 & 0.961 & 0.982 & 0.832 & 0.956\\
 & \textsc{CharacTER}\xspace & 0.990 & 0.986 & 0.950 & 0.963 & 0.981 & 0.775 & 0.978\\
 & \textsc{ITER}\xspace & 0.865 & 0.978 & 0.982 & 0.966 & 0.965 & 0.872 & --\\
 & \textsc{Meteor++}\xspace & -- & -- & -- & -- & -- & -- & --\\
 & \textsc{NIST}\xspace & \textbf{0.997} & 0.984 & 0.980 & 0.944 & 0.988 & 0.870 & 0.944\\
\bottomrule
\end{tabular}
    \caption{Absolute Pearson correlations with human judgments on WMT18 from-English language pairs for 10K hybrid systems. Correlations of metrics not significantly outperformed by any other for that language pair are highlighted in bold. For each language pair, we specify the number of systems.}
    \label{tab:sup-wmt18-from-sys-hybrids}
\end{table*}

\newpage
\begin{table*}[t!]
    \centering
    \gianttablefont
    \begin{tabular}{c|cccccccc}
\toprule
Setting & Metric & cs-en & de-en & et-en & fi-en & ru-en & tr-en & zh-en\\
\midrule
\multirow{13}{*}{Unsupervised}
 & \textsc{BLEU}\xspace & 0.135 & 0.804 & 0.757 & 0.460 & 0.230 & 0.096 & 0.661\\
 & \textsc{CDER}\xspace & 0.162 & 0.795 & 0.764 & 0.493 & 0.234 & 0.087 & 0.660\\
 & \textsc{CharacTER}\xspace & 0.146 & 0.737 & 0.696 & 0.496 & 0.201 & 0.082 & 0.584\\
 & \textsc{ITER}\xspace & 0.152 & 0.814 & 0.746 & 0.474 & 0.234 & 0.100 & 0.673\\
 & \textsc{Meteor++}\xspace & 0.172 & 0.804 & 0.646 & 0.456 & 0.253 & 0.052 & 0.597\\
 & \textsc{NIST}\xspace & 0.136 & 0.802 & 0.739 & 0.469 & 0.228 & 0.135 & 0.665\\
 & \textsc{PER}\xspace & 0.121 & 0.764 & 0.602 & 0.455 & 0.218 & 0.000 & 0.602\\
\bottomrule
\end{tabular}
    \caption{Model selection accuracies (Hits@1) on to-English WMT18 hybrid systems. We report the average of 100K samples and the 0.95 confidence intervals are below $10^{-3}$. We bold the highest numbers for each language pair and direction.}
    \label{tab:sup-wmt18-to-modelselect-hit1}
\end{table*}

\newpage
\begin{table*}[t!]
    \centering
    \gianttablefont
    \begin{tabular}{c|cccccccc}
\toprule
Setting & Metric & cs-en & de-en & et-en & fi-en & ru-en & tr-en & zh-en\\
\midrule
\multirow{13}{*}{Unsupervised}
 & \textsc{BLEU}\xspace & 0.338 & 0.894 & 0.866 & 0.666 & 0.447 & 0.265 & 0.799\\
 & \textsc{CDER}\xspace & 0.362 & 0.890 & 0.870 & 0.689 & 0.451 & 0.256 & 0.799\\
 & \textsc{CharacTER}\xspace & 0.349 & 0.854 & 0.814 & 0.690 & 0.429 & 0.254 & 0.739\\
 & \textsc{ITER}\xspace & 0.356 & 0.901 & 0.856 & 0.676 & 0.454 & 0.278 & 0.811\\
 & \textsc{Meteor++}\xspace & 0.369 & 0.895 & 0.798 & 0.662 & 0.470 & 0.174 & 0.757\\
 & \textsc{NIST}\xspace & 0.338 & 0.894 & 0.857 & 0.672 & 0.446 & 0.323 & 0.803\\
 & \textsc{PER}\xspace & 0.325 & 0.866 & 0.771 & 0.663 & 0.435 & 0.021 & 0.754\\
\bottomrule
\end{tabular}
    \caption{Mean Reciprocal Rank (MRR) of the top metric-rated system on to-English WMT18 hybrid systems. We report the average of 100K samples and the 0.95 confidence intervals are below $10^{-3}$. We bold the highest numbers for each language pair and direction.}
    \label{tab:sup-wmt18-to-modelselect-mrr}
\end{table*}

\newpage
\begin{table*}[t!]
    \centering
    \gianttablefont
    \begin{tabular}{c|cccccccc}
\toprule
Setting & Metric & cs-en & de-en & et-en & fi-en & ru-en & tr-en & zh-en\\
\midrule
\multirow{13}{*}{Unsupervised}
 & \textsc{BLEU}\xspace & 3.85 & 0.45 & 1.01 & 2.17 & 2.34 & 4.48 & 3.19\\
 & \textsc{CDER}\xspace & 3.88 & 0.43 & 0.87 & 1.33 & 2.30 & 4.58 & 3.43\\
 & \textsc{CharacTER}\xspace & 3.77 & 0.49 & 0.94 & 2.07 & 2.25 & 4.07 & 3.37\\
 & \textsc{ITER}\xspace & 3.55 & 0.46 & 1.25 & 1.43 & 4.65 & 3.11 & 2.92\\
 & \textsc{Meteor++}\xspace & 3.70 & 0.41 & 0.69 & 1.13 & 2.28 & 1.40 & 3.50\\
 & \textsc{NIST}\xspace & 3.93 & 0.49 & 1.10 & 1.19 & 2.36 & 1.42 & 3.92\\
 & \textsc{PER}\xspace & 2.02 & 0.46 & 1.71 & 1.49 & 2.25 & 4.22 & 3.20\\
\bottomrule
\end{tabular}
    \caption{Absolute Difference ($\times100$) of the top metric-rated and the top human-rated system on to-English WMT18 hybrid systems. Smaller difference signify higher agreement with human scores. We report the average of 100K samples and the 0.95 confidence intervals are below $10^{-3}$. We bold the lowest numbers for each language pair and direction.}
    \label{tab:sup-wmt18-to-modelselect-diff}
\end{table*}

\newpage
\begin{table*}[t!]
    \centering
    \footnotesize
    \begin{tabular}{c|cccccccc}
\toprule
Setting & Metric & en-cs & en-de & en-et & en-fi & en-ru & en-tr & en-zh\\
\midrule
\multirow{13}{*}{Unsupervised}
 & \textsc{BLEU}\xspace & 0.151 & 0.611 & 0.617 & 0.087 & 0.519 & 0.029 & 0.515\\
 & \textsc{CDER}\xspace & 0.163 & 0.663 & 0.731 & 0.081 & 0.541 & 0.032 & 0.552\\
 & \textsc{CharacTER}\xspace & 0.135 & \textbf{0.737} & 0.639 & \textbf{0.492} & 0.543 & 0.027 & \textbf{0.667}\\
 & \textsc{ITER}\xspace & 0.000 & 0.691 & 0.734 & 0.112 & 0.534 & 0.031 & --\\
 & \textsc{Meteor++}\xspace & -- & -- & -- & -- & -- & -- & --\\
 & \textsc{NIST}\xspace & 0.182 & 0.662 & 0.549 & 0.083 & 0.537 & 0.033 & 0.553\\
 & \textsc{PER}\xspace & 0.179 & 0.555 & 0.454 & 0.062 & 0.535 & 0.032 & 0.539\\
\bottomrule
\end{tabular}
    \caption{Model selection accuracies (Hits@1) on to-English WMT18 hybrid systems. We report the average of 100K samples and the 0.95 confidence intervals are below $10^{-3}$. We bold the highest numbers for each language pair and direction.}
    \label{tab:sup-wmt18-from-modelselect-hit1}
\end{table*}

\newpage
\begin{table*}[t!]
    \centering
    \footnotesize
    \begin{tabular}{c|cccccccc}
\toprule
Setting & Metric & en-cs & en-de & en-et & en-fi & en-ru & en-tr & en-zh\\
\midrule
\multirow{13}{*}{Unsupervised}
 & \textsc{BLEU}\xspace & 0.363 & 0.764 & 0.766 & 0.323 & 0.714 & 0.205 & 0.666\\
 & \textsc{CDER}\xspace & 0.371 & 0.803 & 0.851 & 0.319 & 0.729 & 0.210 & 0.700\\
 & \textsc{CharacTER}\xspace & 0.346 & \textbf{0.853} & 0.781 & \textbf{0.667} & 0.732 & 0.205 & \textbf{0.809}\\
 & \textsc{ITER}\xspace & 0.044 & 0.825 & 0.853 & 0.365 & 0.717 & 0.210 & --\\
 & \textsc{Meteor++}\xspace & -- & -- & -- & -- & -- & -- & --\\
 & \textsc{NIST}\xspace & 0.393 & 0.803 & 0.710 & 0.326 & 0.726 & 0.211 & 0.698\\
 & \textsc{PER}\xspace & 0.387 & 0.719 & 0.624 & 0.301 & 0.725 & 0.211 & 0.678\\
\bottomrule
\end{tabular}
    \caption{Mean Reciprocal Rank (MRR) of the top metric-rated system on to-English WMT18 hybrid systems. We report the average of 100K samples and the 0.95 confidence intervals are below $10^{-3}$. We bold the highest numbers for each language pair and direction.}
    \label{tab:sup-wmt18-from-modelselect-mrr}
\end{table*}

\newpage
\begin{table*}[t!]
    \centering
    \footnotesize
    \begin{tabular}{c|cccccccc}
\toprule
Setting & Metric & en-cs & en-de & en-et & en-fi & en-ru & en-tr & en-zh\\
\midrule
\multirow{13}{*}{Unsupervised}
 & \textsc{BLEU}\xspace & 1.26 & 6.36 & 2.59 & 0.92 & 0.76 & 9.40 & 3.01\\
 & \textsc{CDER}\xspace & 1.25 & 6.70 & 1.90 & 1.41 & 0.87 & 9.37 & 1.75\\
 & \textsc{CharacTER}\xspace & 1.23 & 6.90 & 2.19 & 4.35 & 0.93 & 5.22 & 1.64\\
 & \textsc{ITER}\xspace & 1.25 & 9.14 & 2.52 & 1.52 & 1.35 & 7.33 & --\\
 & \textsc{Meteor++}\xspace & -- & -- & -- & -- & -- & -- & --\\
 & \textsc{NIST}\xspace & 1.24 & 5.28 & 2.55 & 1.02 & 0.75 & 8.82 & 3.34\\
 & \textsc{PER}\xspace & 1.25 & 6.62 & 4.92 & 7.43 & 0.68 & 9.76 & 2.31\\
\bottomrule
\end{tabular}
    \caption{Absolute Difference ($\times100$) of the top metric-rated and the top human-rated system on to-English WMT18 hybrid systems. Smaller difference indicate higher agreement with human scores. We report the average of 100K samples and the 0.95 confidence intervals are below $10^{-3}$. We bold the lowest numbers for each language pair and direction.}
    \label{tab:sup-wmt18-from-modelselect-diff}
\end{table*}

\newpage

\begin{table*}[t!]
    \centering
    \gianttablefont
    \begin{tabular}{ccc}
    \toprule
    Metric & M1 & M2 \\
    \midrule
    \bleu-1 & 0.124$^*$ & 0.135$^*$ \\
    \bleu-2 & 0.037$^*$ & 0.048$^*$ \\
    \bleu-3 & 0.004$^*$ & 0.016$^*$ \\
    \bleu-4 & -0.019$^*$ & -0.005$^*$ \\
    \meteor & 0.606$^*$ & 0.594$^*$ \\
    \rouge-L & 0.090$^*$ & 0.096$^*$ \\
    \metric{CIDEr} & 0.438$^*$ & 0.440$^*$ \\
\bottomrule
\end{tabular}
    \caption{Pearson correlation on the 2015 COCO Captioning Challenge. 
    The M1 and M2 measures are described in Section~\ref{sec:exp}. We bold the best correlating task-specific and task-agnostic metrics in each setting 
    \leic uses images as additional inputs.
    Numbers with $^*$ are cited from \citet{leic}.}
    \label{tab:coco_appendix}
\end{table*}

\newpage 

\begin{table*}[t!]
    \centering
    \gianttablefont
    \begin{tabular}{c|ccc}
    \toprule
    Type & Method & QQP & PAWS\textsubscript{QQP} \\
    \midrule
    \multirow{3}{*}{ \shortstack[l]{Trained on QQP (supervised)} } & DecAtt & 0.939* & 0.263 \\
    & DIIN & 0.952* & 0.324 \\
    & BERT & \textbf{0.963*} & \textbf{0.351} \\
    \midrule
    \multirow{3}{*}{ \shortstack[l]{Trained on QQP + PAWS\textsubscript{QQP} (supervised)}} & DecAtt & - & 0.511 \\
    & DIIN & - & 0.778 \\
    & BERT & - & \textbf{0.831} \\
    \midrule
    \multirow{64}{*}{ \shortstack[l]{Metric (Not trained on QQP or PAWS\textsubscript{QQP})}} 
    & \bleu-1 & 0.737 & 0.402 \\
\bottomrule
\end{tabular}
    \caption{Area under ROC curve (AUC) on QQP and PAWS\textsubscript{QQP} datasets. The scores of trained DecATT~\citep{decatt}, DIIN~\citep{diin}, and fine-tuned BERT are reported by \citet{paws}. We bold the best task-specific and task-agnostic metrics. Numbers with $^*$ are  scores on the held-out test set of QQP.}
    \label{tab:paws_appendix}
\end{table*}

\end{document}